\documentclass[10pt, onecolumn]{article}
\pdfoutput=1
\usepackage[pagenumbers]{cvpr}
\usepackage{graphicx}
\usepackage{amsmath}
\usepackage{amssymb}
\usepackage{booktabs}
\usepackage{float}
\usepackage[pagebackref,breaklinks,colorlinks]{hyperref}
\usepackage[capitalize]{cleveref}
\usepackage{multicol}
\usepackage{pdflscape}		
\usepackage{mathtools}		
\usepackage{setspace}

\usepackage{caption}
\begin{document}
\setlength{\parskip}{0pt}
\onecolumn
\title{SyntEO: Synthetic Data Set Generation for Earth Observation and Deep Learning - Demonstrated for Offshore Wind Farm Detection}

\author{
Preprint, submitted 23. Aug. 2021\\
 \\
Thorsten Hoeser$^{1,*}$ \quad Claudia Kuenzer$^{1,2}$ \\
 \\
$^1$German Remote Sensing Data Center (DFD), German Aerospace Center (DLR) \\$^2$Department of Remote Sensing, Institute of Geography and Geology, University of Wuerzburg\\
 \\
 \small{$^{*}$Corresponding author, \texttt{thorsten.hoeser@dlr.de}}
}

\maketitle

\begin{multicols}{2}
\begin{abstract}
With the emergence of deep learning in the last years, new opportunities arose in Earth observation research. Nevertheless, they also brought with them new challenges. The data-hungry training processes of deep learning models demand large, resource expensive, annotated data sets and partly replaced knowledge-driven approaches so that model behaviour and the final prediction process became a black box. The proposed SyntEO approach enables Earth observation researchers to automatically generate large deep learning ready data sets by merging existing and procedural data. SyntEO does this by including expert knowledge in the data generation process in a highly structured manner to control the automatic image and label generation by employing an ontology. In this way, fully controllable experiment environments are set up, which support insights in the model training on the synthetic data sets. Thus, SyntEO makes the learning process approachable, which is an important cornerstone for explainable machine learning. We demonstrate the SyntEO approach by predicting offshore wind farms in Sentinel-1 images on two of the worlds largest offshore wind energy production sites. The largest generated data set has 90,000 training examples. A basic convolutional neural network for object detection, that is only trained on this synthetic data, confidently detects offshore wind farms by minimising false detections in challenging environments. In addition, four sequential data sets are generated, demonstrating how the SyntEO approach can precisely define the data set structure and influence the training process. SyntEO is thus a hybrid approach that creates an interface between expert knowledge and data-driven image analysis.
\end{abstract}

\section{Introduction}\label{sec:intro}
Over the last decades, Earth observation faced a massive increase in data availability. Today, petabytes of public and commercial remote sensing archives are accessible, which provide data on a global scale with increasing variance in modality. In order to harness this steadily growing amount of data, machine learning became a fundamental toolset in Earth observation. Lately, with the emergence of deep learning and especially its development in computer vision for image analysis \cite{lecun2015}, it became possible to extract complex spatio-temporal features from large amounts of remote sensing data. As a consequence, deep learning eventually established itself as an essential tool in Earth observation and gave new perspectives to geoscientific research \cite{zhu2017deep, reichstein2019deep, hoeser2020II, tuia2021agenda}.

The successful implementation of deep learning at a time of increasing data availability, not just in Earth observation, is no coincidence but rather a self-reinforcing effect. Deep learning needs huge and precisely labelled training data in order to learn the underlying complex features. At the same time, a deep learning model is highly efficient in processing large amounts of unseen data once trained. Hence, deep learning is both an approach that is capable of processing large data archives as well as an approach that highly depends on large training data sets \cite{krizhevsky2012, lecun2015}.

The latter brings up a major drawback of deep learning: The creation of large, precisely labelled training data sets is a resource-expensive task. Multiple approaches have been developed to cope with the problem that deep learning demands large annotated training data sets. Few-shot learning is a meta-learning approach that trains a model to perform a task, such as a general separation of classes instead of predicting a specific class. That way, a large training data set can be used to train a task which can then be applied to a small support set of target data which alone would be too small to optimise a deep learning model. However, an initial large training data set, prior knowledge or a combination of both is needed \cite{sun2021research}. In weakly supervised learning, large data sets are used, which have limited labels like inaccurate labels or labels with high-level information compared to the task at hand. That way, annotating data is less time consuming, and more data can be acquired at the same time or even be automatically generated. During the training process, limited labels are refined iteratively to receive sufficiently labelled training examples. That way, tasks can be learned on limited labels, which from a \textit{strong} supervised perspective would not be sufficient \cite{zhang2015weakly}. In self-supervised learning, models start learning representations in a data set without using a single label. Instead, they produce their own supervision signal during training. To do so, so-called pretext-tasks have to be defined, on which the model can generate its own supervision signal. For example, such a task can be to remove noise from an automatically generated noisy image, where the image is the raw training data and the noise generated on the fly and therewith known to the model. By solving this pretext-task, the model learns representations within the data set. After that, the model can be trained on the target task, like object detection with a smaller amount of labelled data \cite{tao2022remote, akiva2021self}.

All of these approaches have one thing in common. At some point, they need a large amount of unlabelled or imperfect labelled images, which can become a problem in the Earth observation domain. Depending on the sensor and target of interest, even today, the number of usable images can quickly get too small to optimise a deep learning model. Therefore, we focus on synthetic training data in this study, where images and labels are generated automatically. Thus, the raw data limitation and annotation problems are solved simultaneously, and the commonly employed supervised learning approach can still be used without adapting well-established training schemes.

The successful application of synthetic generated data sets in other domains like computer vision, bioinformatics, or robotics demonstrate the potential of this approach in combination with deep learning \cite{nikolenko2021synthetic}. However, in Earth observation, the number of studies that incorporate synthetic data into their deep learning workflows are vanishingly small compared to the surge of publications that apply deep learning methods in the last years \cite{hoeser2020II}. In addition to generating large training data sets, synthetic data opens up new opportunities towards a sophisticated human-machine interaction that supports an understanding of the model's reasoning. The latter is part of the recently growing eXplainable AI (XAI) field, intending to open the black box of artificial agents to make models interpretable \cite{gunning2019xai, Gunning2019darpa}.

\begin{figure}[H]
\centering
\includegraphics[width=15.5 cm]{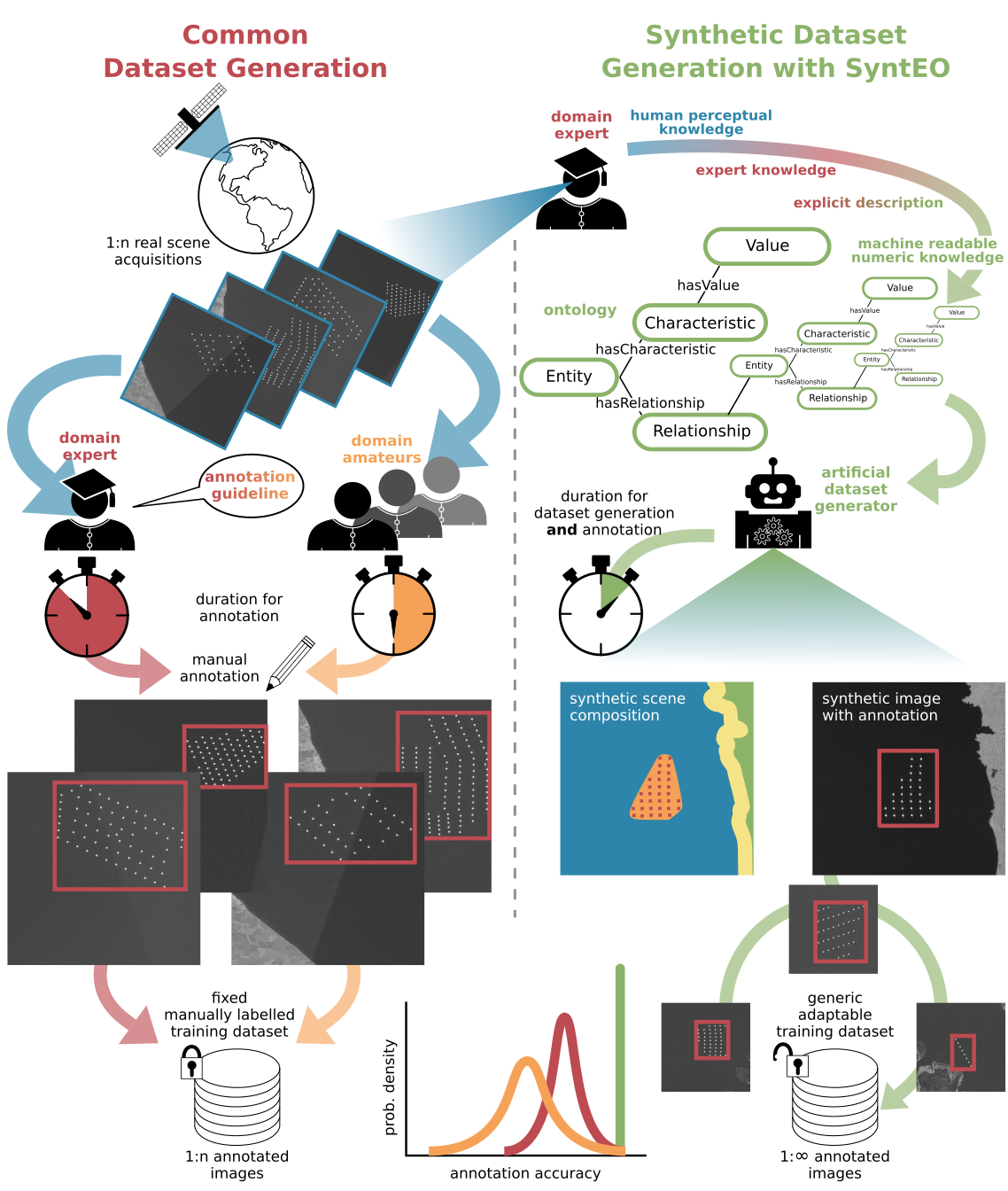}
\caption{Comparison of two data set creation processes. Left: The common way of manual annotation which creates fixed and limited data sets. Right: The proposed approach of synthetic data set generation with fully automatic annotation and dynamic data sets in size and content.}
\label{fig:common_synteo}
\end{figure}

In order to promote the usage of synthetic data sets in Earth observation, we introduce SyntEO, an approach for synthetic data set generation with a particular focus on data from the Earth observation domain. Figure \ref{fig:common_synteo} summarises how a deep learning data set is commonly created and compares it to a process that applies SyntEO. In common data set creation, a domain expert visually examines a limited set of sensor acquisitions and then annotates the images by hand. Each image-annotation pair is then stored as a training example, which results in a limited data set with as many training examples as acquisitions are available. The sole annotation by a domain expert occupies the most resources but will result in a relatively high annotation accuracy. To redistribute the workload and free resources, the domain expert can formulate an annotation guideline to instruct domain amateurs if the complexity of data annotated and topic allows this knowledge transfer. The resulting data set annotated by the domain amateurs will be technically the same, a limited amount of image-annotation pairs. However, annotation accuracy might be lower with a wider variance than the same data set only labelled by a single domain expert.

On the right-hand side of figure \ref{fig:common_synteo} a synthetic data set generation process that uses the SyntEO approach is depicted. A domain expert has to explicitly formulate machine-readable numeric knowledge, which describes how the domain expert perceives characteristics from single entities and their interrelations in the data. This is done by formulating an ontology (see section \ref{sec:onto}) that provides dimensions of numeric knowledge for the observed characteristics. Upon this ontology, an artificial data set generator can interpret the knowledge representation and compose a synthetic scene of all entities described by the expert. A synthetic image is generated by adding texture to the composed scene, whereas the synthetic scene composition can be used to derive different types of annotations that will always be accurate without any variation. Nevertheless, the quality of the represented task and remote sensing scene depends on the expert knowledge in the ontology. Since the artificial generator relies on a complex description of potential characteristics and randomised value selection, the number of possible training examples is to be assumed infinite.

Therewith, SyntEO has two major goals:
\begin{itemize}
\item To enable domain experts to create large deep learning ready data sets specifically designed for Earth observation research and remote sensing data by automatic image and annotation generation.
\item To provide an interface that can be used for human-machine interaction, where an initial change in parameters during the data generation process triggers different behaviours in the machine learning model, which learns from the generated data. That way, an interactive loop is established where the user and model react to each other's outputs.
\end{itemize}

\end{multicols}
\begin{multicols}{2}

In this introduction of the approach, first, the theoretical background is presented by discussing the general motivation and core concepts of SyntEO. Following that, the first application of SyntEO for offshore wind farm detection in Sentinel-1 imagery demonstrates a hands-on example for the efficient creation of a large data set to solve a real-world problem. This example gives an intuition on how SyntEO provides an experimental environment to get insights into the learning process and model behaviour and how to adapt the data set creation process to accomplish accurate, spatially transferable offshore wind farm detection by minimised false detections.

\section{Related Research}

\subsection{Synthetic Data sets in Earth Observation}

Especially in computer vision, synthetic data sets are an established tool to train neural networks for tasks like predicting optical flow \cite{dosovitskiy2015flyingchairs}, object detection \cite{tremblay2018fallingthings} or image segmentation \cite{khan2019procsy}. To create these data sets, different strategies have been used, like adding virtual models of objects in front of random backgrounds \cite{dosovitskiy2015flyingchairs} up to generating entire 3D environments and using virtual cameras to take pictures of them \cite{khan2019procsy}. Thus, synthetic training data can be seen as a recomposition of already existing template data as well as fully synthetically generated data.

In Earth observation, the number of studies that combine synthetic data sets and deep learning is a rarity despite its successful implementation in other research domains. Of those studies which combine a synthetic data set and deep learning approach, RGB images of very high resolution with less than 1~m are the majority. Their applications are aircraft or vehicle detection \cite{han2017efficient, berkson2019synthetic, shermeyer2021rareplanes, weber2021artificial} and building footprint detection \cite{kong2020synthinel1}. Two more synthetic data sets focus on speckle noise reduction in radar imagery \cite{dabhi2020virtual, vitale2021analysis} and one publication uses fully synthetic training data for river network extraction from single band MNDWI images of Landsat~8 acquisitions \cite{isikdogan2018learning}.

For data set generation Isikdogan~\etal \cite{isikdogan2018learning} and Kong~\etal \cite{kong2020synthinel1} used fully synthetic approaches. Especially Kong~\etal \cite{kong2020synthinel1} preserved spatial context accurately by generating complete 3D environments in which objects are located not randomly but in spatial relationships similar to real-world environments. In contrast, Weber~\etal \cite{weber2021artificial} demonstrated that 2D models of vehicles that were placed in front of random backgrounds and therewith dismissing spatial context are still detectable in real-world overhead images with a very high spatial resolution of 10~cm. However, since such feature richness of the object of interest is not guaranteed across application domains and especially spaceborne sensor spatial resolution, we argue that spatio-temporal context of target objects and their environment is a major characteristic of Earth observation data and has to be an important aspect in synthetic data generation.

One common feature shared by all the studies mentioned is the training of the convolutional neural network (CNN) deep learning model with the synthetically generated data. Since CNNs are the most widely used deep learning models in Earth observation \cite{hoeser2020II, ma2019deep} and also used in this study, we briefly outline the development and characteristics of CNNs.

\subsection{Convolutional Neural Networks in Earth Observation}

When in 2012 CNNs became popular with the introduction of AlexNet, stacked convolutional layers with trained kernel functions were used to predict single labels for entire images on relatively small inputs with a dimension of $224\times224$ pixels \cite{krizhevsky2012}. The so-called patch-based approach used the same general architecture designs with even smaller input sizes for image segmentation or object detection. In order to find objects or to segment an image, the image was split into patches of $8\times8$ to $16\times16$ pixels. For each patch, a prediction was made if the patch or patch centre pixel shows an object or belongs to a semantic class. The aggregation of all predictions on each patch was then the result for a given image. It is important to mention that at this early stage in the evolution of CNNs, the small spatial range of $8\times8$ to $16\times16$ pixels can only extract limited spatial features. In an Earth observation context, this limitation was interpreted to be close to a pixel-based approach because single pixel information still dominates features generated from such spatially small patches \cite{lang2019geobia}. However, convolutional feature extractors have developed fast and at least since the introduction of the Fully Convolutional Network (FCN) for image segmentation \cite{shelhamer2014fully} or Region based-CNN (R-CNN), especially Faster R-CNN for object detection \cite{ren2015faster}, new end to end trainable CNN architectures with considerable larger input sizes became available. With the development of graphics processing units (GPUs) with increased memory, the input size and the capability of CNNs to learn large-scale spatial features from the training data became more sophisticated. An important reaction to this development was the creation of new building blocks for CNN architectures. These new architectures specifically search for spatial features on multiple scales in order to combine small and large scale features \cite{chen2017rethinking, lin2016feature}.

Overall, object detection and image segmentation with CNNs changed significantly over the last decade and are today capable of extracting features from large input sizes on multiple scales by taking spatial context into account. This development is especially important for the successful application in Earth observation since fine-grained features on multiple scales are essential characteristics of remote sensing data. For an in-depth review of the evolution and application of CNNs in the context of Earth observation, we refer to Hoeser and Kuenzer \cite{hoeser2020I} and Hoeser~\etal \cite{hoeser2020II}.

\subsection{Human-Machine Interaction}

With the rise of artificial intelligence in the last decades, humans have gotten used to interacting with machines daily. Smartphones guide our cameras by recognising faces and providing optimal settings that the user can see and select. Digital assistants like Alexa and Siri understand questions and commands due to their speech recognition modules and perform tasks like turning lights on or buying a product. Humans are getting used to these intelligent agents and, to some extent, trust that they will perform the task for which they were designed without knowing how they were designed.

However, in science, it is of great importance to interact with models and understand how they make their decisions. When a user applies a decision tree based on expert knowledge to analyse a remote sensing scene, the outcome can be understood in its entirety since the model is fully interpretable. Instead, deep learning models like the CNN are categorised as black-box models due to the vast amount of parameters. A user who applies such a model typically accepts the output of the model without knowing how the model came up with the prediction \cite{arrieta2020explainable}. Thus, using a standard CNN, there is limited interaction between the user and the artificial agent here a machine learning model. The user requests a prediction, which the artificial agent delivers, but the user can not request an understandable explanation.

Metrics that communicate the model performance are a first step to building confidence in the prediction. However, they do not explain how the model works, and a model can successfully solve a task, but it might do so based on completely different features than those assumed by the user \cite{das2020opportunities}. This can be problematic since a user who thinks to understand the reasoning of the model might draw wrong conclusions from the model output \cite{Gunning2019darpa}. Also, the model's competence is unknown to the user. That means that the model's performance can drop drastically when it operates in a new situation. However, the user does not necessarily know what a new situation looks like to the model. Therefore, knowing a models competence is of high value since a user can better decide when the limits of the model are reached, and the model needs adaptations or, if adaptations are not possible, stop relying on it \cite{gunning2019xai}.

Since deep learning based artificial agents are challenging to explain, let alone make them completely interpretable \cite{arrieta2020explainable}, the possibility to increase interaction with the model in fully controllable experiment environments can help to formulate a hypothesis about their reasoning and competence. Moreover, increasing interaction and gaining insights into model behaviour can be used to adapt models to enhance their competencies. This application of eXplainable AI~(XAI) is described as "explain to control" by Adadi and Berrada \cite{adadi2018peeking}~p.~(52143). With the emergence of artificial agents over the last decade in science, industry, healthcare, security and day to day life, XAI gains increasing importance \cite{gunning2019xai, Gunning2019darpa} and was lately promoted to be of high interest in Earth observation \cite{tuia2021agenda}.

\section{Introduction to SyntEO}

Considering the limited amount of studies that apply synthetic data in Earth observation with deep learning, a guideline is necessary to encourage researchers to explore the opportunities of synthetic training data. Such a workflow should enable users to automatically generate large amounts of labelled training data upon their expert knowledge, existing template data, and newly generated data. During the data generation process, spatio-temporal context on multiple scales should be considered to simulate the inherent characteristics of Earth observation data. Furthermore, the workflow should support to add and modify parameters in order to expand the complexity of the synthetic data sets. Thereby, stable experiment environments are established that researchers can use to explore which changes in the data set influence the training process and model behaviour.

To that end, we introduce SyntEO, a synthetic data set generation approach specifically designed for investigations of Earth observation data and deep learning methods. To make this introduction more intuitive, we will use examples of the later investigated offshore wind farm detection to illustrate our explanations. Nevertheless, the SyntEO approach is by itself a generic approach, and the discussed examples are only chosen to explain the otherwise abstract and technical SyntEO workflow.

\begin{figure}[H]
	\centering
	\includegraphics[width=16 cm]{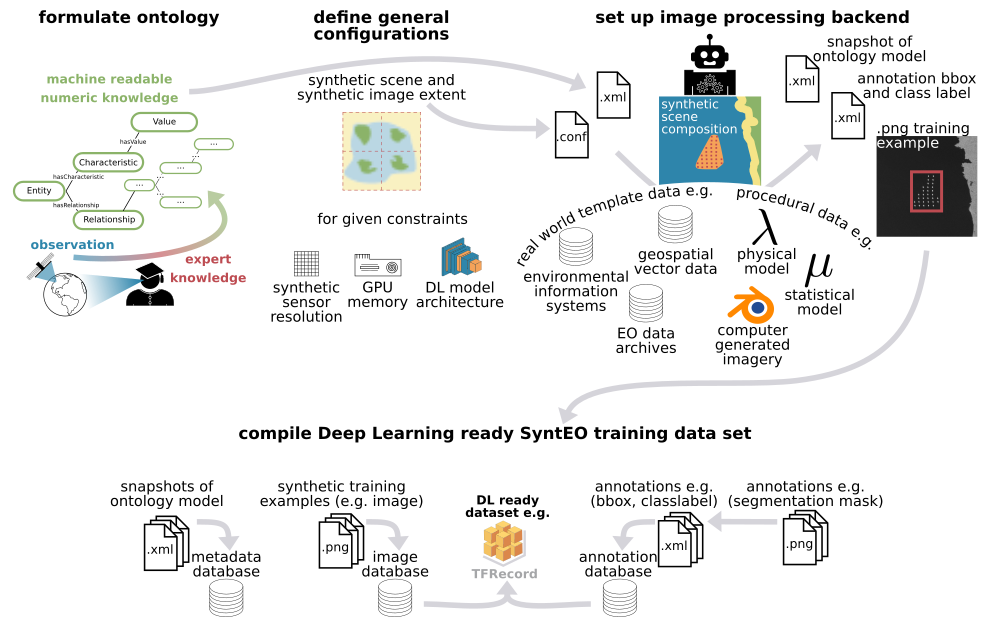}
	\caption{Overview of the SyntEO workflow. First, the domain expert formulates a machine-readable ontology and defines general configurations. Then, an artificial data set generator uses this information to generate a complex data set from template and procedural data. The resulting image-annotation pairs are then transformed into deep learning ready data sets.}
	\label{fig:synteo_workflow}
\end{figure}

The SyntEO workflow consists of four steps pictured in figure \ref{fig:synteo_workflow}. First, an expert formulates an ontology in which perceptual knowledge from observing real-world examples and remote sensing data as well as expert knowledge is formalised to become machine-readable, numeric knowledge. Using an ontology, human-understandable semantic in the knowledge description is preserved and accessible as machine-readable numeric knowledge and logics. To formalise expert knowledge about offshore wind farms in the SyntEO ontology, an Earth observation expert would start by describing the outer boundary as \textit{large} or \textit{small} polygons with specific shape parameters along with numerical values in square meters and number of edges. Likewise, the internal structure of how wind turbines are distributed in these bounds is described and finally, a single turbine. Thus, the ontology holds nested information of entities related to each other.

The second step defines general configurations for the data generation process. Hereby, the synthetically generated training data dimensions are defined as the synthetic scene and image extent. They are important to solve trade-offs between characteristics of the spatial extent of the target, the granularity of characteristic features and constraints given by hardware capacities or the deep learning model architecture. When this trade-off is solved to maximise potential feature representation in the synthetic training data, the machine-readable ontology and the general configurations are passed to an artificial data generator. An example configuration would take into account that an entire offshore wind farm appears on a large spatial scale but has small scale information represented by single wind turbines simultaneously. Thus, a scene and image extent have to be found, in which the wind farm entity can appear entirely at a pixel resolution that also captures small scale features of single turbines without exceeding the memory of the hardware, which is later used for training a deep learning model.

\end{multicols}
\begin{multicols}{2}

The artificial data generator uses the synthetic scene extent to initiate boundaries of a virtual environment. By following the ontology, values are selected, which describe characteristics of entities to generate geometries of single scene elements and finally compose them to a two or three-dimensional discrete description of a synthetic scene. The single geometries in this composition are then filled with class-specific texture to simulate a sensor acquisition and derive task-specific annotations. The artificial data generator can be imagined as a cluster of user-defined modules that can query databases to add existing template data to the scene composition or generate fully synthetic data, which we call procedural data.

The last step is the export of the generated data to a training example like an image and annotation pair and a metadata file that describes the selected values of the ontology by the artificial data generator for each example. All generated training examples are summarised in a database and finally optimised as a deep learning ready data set.

The initial motivation of the SyntEO workflow is to systematically generate Earth observation data that simulates spatio-temporal features and context. Technically this is done by the artificial data generator, which employs the ontology in order to generate a contextual scene composition of hierarchically nested targets and non-targets on multiple scales \cite{wu1995hierarchy}. Therewith, SyntEO is based on the assumption that space and spatial features are inherent characteristics of Earth observation data. For spatial features to appear in Earth observation data, it is necessary that the sensed objects are significantly larger than the spatial resolution of the sensor. This relation is described by the H-resolution remote sensing scene model proposed by Strahler~\etal \cite{strahler1986on}. A remote sensing scene is described by Strahler~\etal \cite{strahler1986on} "as the spatial and temporal distribution of matter and energy fluxes […]" (p.~122) and furthermore to be "[…] not chaotic or random, but manifest spatial and temporal order" (p.~123). Derived from this, a synthetic scene in SyntEO is the spatial and temporal non-chaotic composition of scene elements which are an abstract representation of matter and energy fluxes. Hereby, scene elements can be two or three-dimensional targets and non-targets that describe the spatial location, size, distribution and shape of, e.g. land cover, land use classes, atmospheric conditions, natural or artificial landscape and objects. Targets and non-targets are both scene elements. Thereby, a target is an element of interest that is to be predicted by a model. Whereas non-targets are scene elements that naturally appear or specifically do not appear within a spatial range of targets. In particular, the term background is not used in SyntEO since we argue that a remote sensing scene is a composition of spatially distributed information that as entirety can communicate more features which help to make accurate predictions on a target due to their spatial arrangement together with non-targets compared to a target in front of a random background.

\subsection{Ontology Formulation}\label{sec:onto}

\end{multicols}
\begin{figure}[H]
\centering
\includegraphics[width=16 cm]{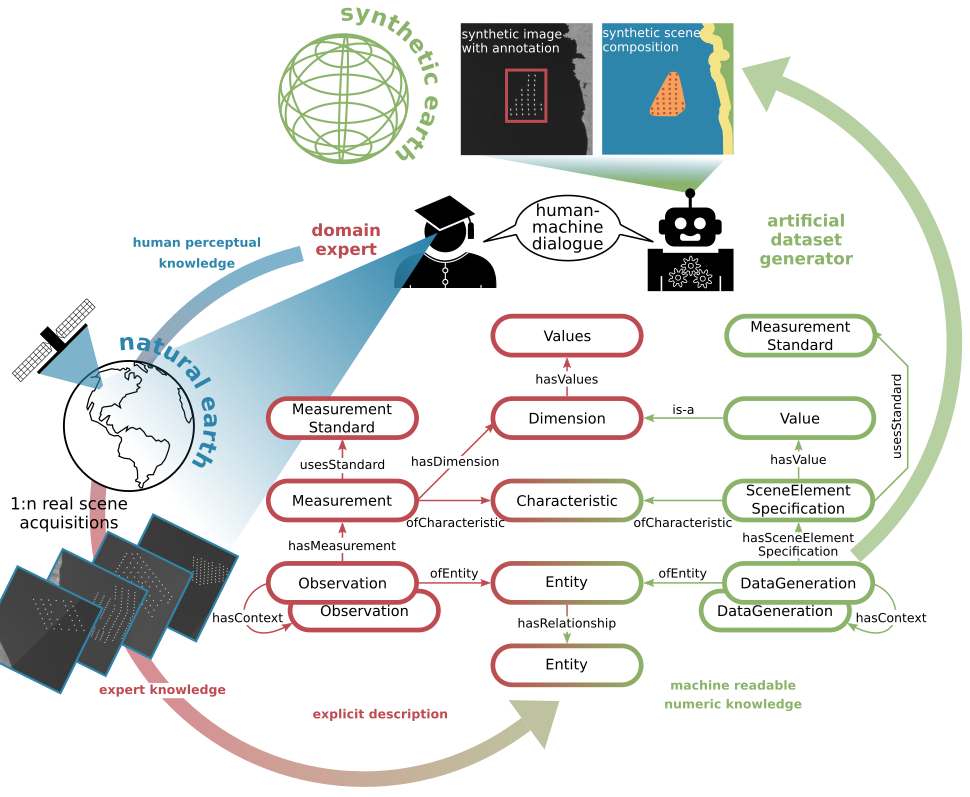}
\caption{Overview of the core classes in the SyntEO ontology. Red classes are used by the domain expert to include knowledge starting at an \textit{Observation} in the ontology. Green classes belong to the artificial data generator, which ingests the represented knowledge in the ontology by accessing it at the \textit{DataGeneration} class to create training examples. Red and green classes are shared by both sides to establish a human-machine dialogue.}
\label{fig:synteo_ontology}
\end{figure}
\begin{multicols}{2}

In order to automatically generate training data with nested, multi-scale scene elements and compose them to a harmonic, synthetic scene that takes spatio-temporal context into account, a well-structured description of properties has to be used as a general basis. Furthermore, this description has to be approachable by a human expert as well as an artificial data generator. The expert must be able to include expert knowledge that describes the properties of single scene elements and their relationships. At the same time, the artificial data generator must be able to ingest this information to generate synthetic training data. Therefore, semantics and numeric and logical expressions are essential to this structured knowledge representation. For these reasons, an ontology is used in the SyntEO approach.

In SyntEO, an ontology is defined from an artificial intelligence and knowledge engineering perspective to be a formal, non-ambiguous representation of knowledge \cite{agrawal2005ontological}. More specifically, an ontology in SyntEO stores information about nested entities which together describe the properties of a remote sensing scene. The goal is to make perceptual and expert knowledge explicit, structured and finally machine-readable. Therewith, an ontology in SyntEO follows the definition of Gruber \cite{gruber1995toward} to be a "simplified view of the world that we wish to represent for some purpose" (p.~908). Where in SyntEO, the purpose is to create synthetic data upon this ontology generically. Ontologies have already been used in GIScience and remote sensing, often for the purpose of adding semantics to large geo- or remote sensing databases \cite{ciodescu2011osmonto, amiri2017fuzzy} or directly for image analysis \cite{nasri2018towards, moran2017combining}. For an in-depth introduction to ontologies and their applications in GIScience, Earth observation and ecology, we refer to Agrawal \cite{agrawal2005ontological}, Arvor~\etal \cite{arvor2019ontologies} and Madin~\etal \cite{madin2007oboe}.

The ontology in SyntEO can be approached from two perspectives: A domain expert who defines entities and their characteristics, and the artificial data generator which uses this description to generate synthetic examples. Thereby, the domain expert's perspective has to be a description of dimensions in which the characteristics of entities can possibly appear. In contrast, the data generator selects single values of these dimensions to generate unique compositions of entities, resulting in synthetic training examples. Therefore, it is important that the dimensions of entity characteristics are related to guarantee coordinated value selection by the data generator.

The Extensible Observation Ontology (OBOE) \cite{madin2007oboe} was chosen as starting point and adapted for the SyntEO approach. Figure \ref{fig:synteo_ontology} shows the core classes and scheme of a single entity description in a typical SyntEO ontology. The domain expert’s perspective, depicted in red, starts at an \textit{Observation} of an \textit{Entity}. Each \textit{Observation} can have multiple \textit{Measurements} where each \textit{Measurement} is of a specific \textit{Characteristic}. Since the domain expert has to describe \textit{Dimensions} instead of single values, each \textit{Measurement} has a \textit{Dimension} which again has \textit{Values} which describe the \textit{Characteristic}. The \textit{Values} in the \textit{Dimension} can take on a wide variety of formats like, statistical distributions with limiting ranges, a set of possible predefined choices or template information coming from linked databases.

To better illustrate this scheme, let us assume the formal description of the size of offshore wind farms (OWF). First, a domain expert makes \textit{Observations} of the \textit{Entity} offshore wind farm. The \textit{Measurement} of \textit{Characteristic} ‘size’ has a \textit{Dimension} which is then described by the domain expert with three predefined possible choices with the \textit{Values} ‘small:5 km’, ‘medium:10 km’ and ‘large:17 km’. The additional usage of the vague terms (small, medium and large) is an example of how to make numeric knowledge better understandable for a human interpreter.

After the domain expert has included this information in the ontology and made it explicit, the artificial data generator uses it for its purpose of \textit{DataGeneration}. For each \textit{DataGeneration} of an \textit{Entity} and its \textit{Characteristics} the data generator selects a \textit{SceneElementSpecification}. This single \textit{Value} originates from the \textit{Values} in the \textit{Dimension} which were defined by the domain expert. In case of the OWF size, the artificial data generator would start a \textit{DataGeneration} for a single OWF and its \textit{SceneElementSpecification ofCharacteristic} ‘size’. From the predefined \textit{Values} in the \textit{Dimension}, it then selects a single \textit{Value} e.g. ‘small:5 km’.

Since the artificial data generator has to harmonise all selected single scene elements and their \textit{SceneElementSpecifications} to become a coherent scene composition, \textit{Context} and \textit{Relationships} are added. \textit{Context} is used to link \textit{Observations} and \textit{DataGeneration} of different \textit{Entities} as reactive bindings in nested hierarchies. For example, a ‘small:5 km’ OWF size causes a higher wind turbine density, which properties are described in an other \textit{Entity} that is linked via the \textit{Context} attribution. The class \textit{Relationship} represents the spatio-temporal connections between \textit{Entities}. Here, the \textit{Relationship ofType Topology} are used to represent spatial conditions, for example, a wind turbine \textit{Entity MustBeCoincidentWith} an offshore wind farm \textit{Entity} which again \textit{MustNotOverlap} with a land \textit{Entity}. By implementing \textit{Context} and \textit{Relationships}, a nested structure can be represented in the ontology and from an artificial perspective \textit{Entities} can now be handled as sets. For the given \textit{Relationship} example the transitive relation can be derived that an off shore wind turbine can not appear on land, without that the \textit{Entities} of off shore wind turbine and land were directly connected in the ontology.

\subsection{Synthetic Scene and Image Extent}

General configurations must be defined before the artificial data generator can utilise the ontology for composing a synthetic scene. These configurations describe the extent of the synthetic environment in which the synthetic scene is composed as well as the extent and resolution of the synthetic image which is generated. The synthetic scene extent is important in order to initialise a data generation environment that is aligned to the information within the ontology about the extent of single scene elements and their relationships. The synthetic image extent and the synthetic sensor resolution control the maximum size and granularity of the features represented in the image. Especially by defining the synthetic image extent, the available hardware and deep learning model architecture are to be considered. Finally, synthetic scene and image extent have to be coordinated with each other. In order to define optimal synthetic scene and image extent, the following aspects have to be considered:

\begin{itemize}
\item The spatial or temporal extent of potential features which can be derived from the extent of single scene elements and their relationships.
\item The synthetic sensor resolution which will finally represent the feature's occurrence in the training data. Thereby, the synthetic sensor resolution should be the same as of the sensor that acquires the data which the trained model will predict on in production.
\item The deep learning model architecture.
\item The specification of the hardware which will be used for training.
\end{itemize}

For a better intuition of synthetic scene and synthetic image extent, let us consider the following example. A synthetic scene has a maximum extent which the creator defines. This extent can be infinite or finite, whereas the maximum synthetic image extent is always finite. Hence, the scene extent can be the same size or larger as the maximum image extent. Thus, from a composed, single synthetic scene, one or multiple images can be generated. For the sake of simplicity in this introduction, the maximum scene extent is considered to be equal to the maximum image extent.
\end{multicols}
\begin{figure}
\centering
\includegraphics[width=12 cm]{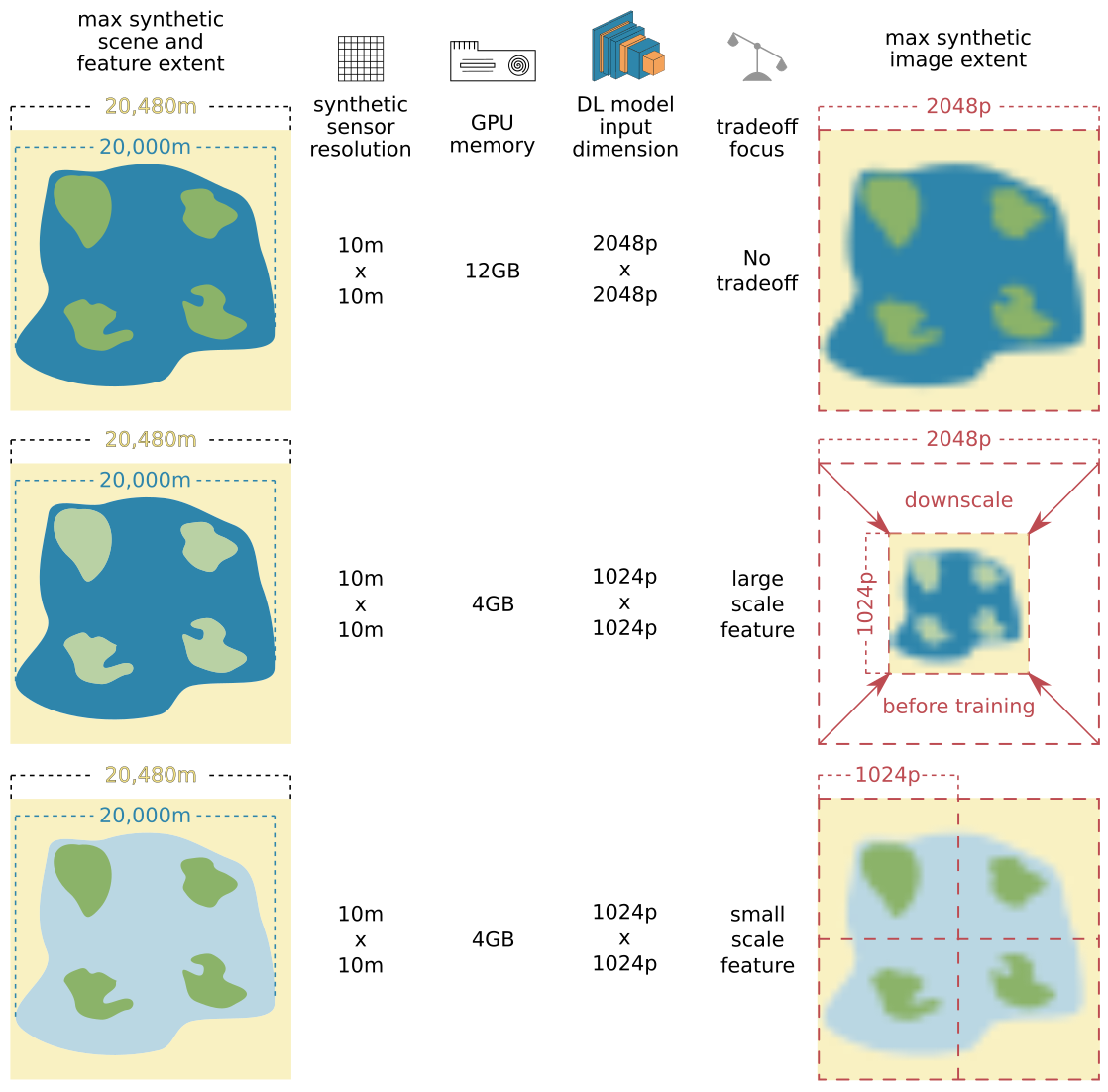}
\caption{Three examples of how to define the maximum virtual scene and image extent under given constraints in feature extent, virtual sensor resolution, hardware and DL model architecture. The constraints for the first example allow using the full synthetic scene at full sensor resolution, that no trade-off between small or large features is necessary. In examples two and three, the given constraints demand smaller image dimensions, which in (2) leads to downscaling of the full image and loss in fine-grained features. In (3), multiple images are taken from the same scene to maintain small features at the cost of large features in the training image.}
\label{fig:scene_ext}
\end{figure}
\begin{multicols}{2}
In order to define the maximum scene extent, the creator must decide on which spatial scale the largest feature can occur potentially. Figure \ref{fig:scene_ext} shows an example that assumes a maximum feature extent (depicted as a blue polygon) of 20~km $\times$~20 km and a synthetic sensor resolution of 10~m $\times$ 10~m. The following three examples show how trade-offs can be balanced in order to maximise feature representation in the synthetic image. Since maximum feature extent defines the space in which a scene must be composed, the maximum scene extent is aligned to the largest potential feature size. Because the final result is an image, the synthetic sensor resolution is considered to refine the scene extent to a technically suitable value. Here, maximum feature extent and synthetic sensor resolution would result in an image of 2000 $\times$ 2000 pixels. Since $2^n$ are common dimensions in CNNs, the final synthetic scene extent is adjusted to be 20,480~m $\times$ 20,480~m so that an image of 2048 $\times$ 2048 pixels can be taken of that synthetic scene.

Looking at figure \ref{fig:scene_ext}, one can see that changes in GPU memory and deep learning architecture can lead to situations in which the entire synthetic scene can not appear in a single synthetic image by maintaining the highest resolution. The first example shows a situation where GPU memory is large enough to optimise a deep learning model with an input size of 2048 $\times$ 2048 pixels. Thus, no trade-off has to be balanced, and the synthetic scene with features of large and small sizes can appear in original sensor resolution in a single synthetic image. In examples two and three, the trade-off balancing is discussed in the context of a decrease in available GPU memory, which forces the deep learning model to use a smaller input dimension of 1024 $\times$ 1024 pixels to avoid a memory overflow during model optimisation. Note that the GPU memory decrease is exemplary. Each of the other aspects can also induce the need to balance trade-offs. For scenario two, the domain expert decides that large scale features (blue polygon) are more important than small scale features (green polygons). In order to ensure that large scale features appear entirely in a single training example, a synthetic image with 10~m $\times$ 10~m resolution and a size of 2048 $\times$ 2048 pixels is first taken from the synthetic scene. Before training, this image is downscaled to the input size of 1024 $\times$ 1024 pixels. The downsizing maintains the appearance of the large feature in one image. However, since the pixel size is artificially decreased to 20 m $\times$ 20~m, a small-scale feature representation loss has to be accepted. In the third example, the domain expert decides that small scale features are more important than large scale features. Thus, the higher spatial sensor resolution of 10~m $\times$ 10~m has to be unchanged to ensure that fine-grained features appear. Four synthetic images of size 1024 $\times$ 1024 pixels are taken from one synthetic scene, splitting the large scale feature and downgrading its representation in the training data.

In all of the three scenarios, the synthetic scene remains the same. This is important since it allows to compose a synthetic scene coherently for small and large scale features, even when the trade-off focus later shifts to e.g. small scale features. That way, both features still appear together in a meaningful relationship in the synthetic image, even when the larger scene element is not entirely included in the final synthetic image but in the synthetic scene.

\subsection{Artificial Data Generator}

The artificial data generator can create synthetic training examples with the formulated ontology and defined general configurations. In a first step, a discrete scene composition is generated within a predefined synthetic environment which describes where or when data of a specific scene element has to appear in relation to other scene elements. The texture is added to the discrete geometries to simulate the sensor's measurements in a second step.

The data generator can be understood as a cluster of modules connected to the ontology that either query and rearrange existing data or process completely new synthetic data. Thereby, each module uses the parameters which are provided by the ontology. The way the information is stored in the ontology can be different for each scene element and if it is a geometry or texture. For example, vector databases like OpenStreetMap (OSM) can be queried and selected by the artificial data generator to receive geometries for specific scene elements. The same is possible for texture upon the predefined geometry. An Earth observation archive can be searched, and the raster data included for that specific scene element. Therefore, the databases are linked in the ontology as \textit{Values} of the \textit{Entity's Dimension}. In SyntEO we refer to data within \textit{Dimensions} which already exists as \textit{template data}. On the opposite are fully synthetic generation processes, in which geometries and texture are generated procedurally. Such data is referred to as \textit{procedural data}. For instance, processes from the computer graphics domain and their implementation in free tools such as Blender \cite{blender2021blender} can be connected to the ontology as modules that generate fully synthetic geometries and texture. However, more specialised physical and statistical models which describe scene elements in specific research domains and sensor specifications can also be plugged into the data generation process. Tools like RaySAR \cite{auer2016raysar}, or spectral libraries \cite{baldridge2009aster} are then used to generate the data described by the ontology. The modular design of the data generator and its close interaction with the ontology enables to link between existing data archives and existing synthetic approaches and models and novel generator modules tailored to generate scene elements for specific research projects.

After the last step in the data generation process, a virtual image is taken from the generated data when the texture was added to the scene composition. Together with the synthetic image, the necessary annotation is extracted from the final scene composition. The discrete geometries provide ground truth information in the same depth as formulated in the ontology and deriving segmentation masks or bounding boxes is easily possible.

In this example, the generated data, a 2D image, and annotation are both saved to separate files. To support efficient access to the training examples, translating the data to binary file formats like the TFRecord format for the TensorFlow deep learning framework is optional but recommended. In addition to the training data, the SyntEO approach allows saving a snapshot of the sampled ontology from the data generator perspective as a detailed description of all parameters which were used to generate the scene elements and their composition. The accumulation of all snapshots is an important source for a controllable experiment environment and a strong tool in ablation studies to gain insights into the model training. The default file format for an ontology snapshot is the same as for the origin ontology, a .xml file.

\subsection{Wrap-up and Potential Applications}

For simplicity, SyntEO can be reduced to two core functionalities: The ontology formulation and the artificial data generator. The formulated ontology can be seen as a complex parameter file with nested information which describes a remote sensing scene from the big picture to tiny details. The artificial data generator is the computational backend that ingests this description and builds images based on it. The ontology can be understood by humans and machines and maps expert knowledge to technical modules that compose the synthetic remote sensing scenes. The technical interface between domain experts and artificial data generator can be a .xml or .json file or any format, which supports to store information of nested entities.

The image processing backend has to realise the description embedded in the ontology. For example a simple ontology could describe a single band binary image with a circle of random radius at the centre of a raster of a specific extent. A straightforward processing backend can be a python script, which constructs a circle in vector space and uses the provided raster dimension to rasterise this circle. The final output would then be a binary image showing the circle and a label which, for example, holds the circle radius information. The upcoming hands on example of offshore wind farm extraction is more complex. Here the processing back end are mainly two modules: A remote sensing image database from which texture examples are queried and a python script that uses 2D numpy arrays to construct geometries of specific shapes and create procedural texture information. Together, they use the parameter file defined by the ontology to create single entities independently before they are combined, again by following the description in the parameter file.

Since the synthetically generated radar images will be a single band image with a 10~m resolution, the open question is if SyntEO can be used for more complex data like multi-channel images with a very high resolution. The following example should provide an intuition about the motivation and transferability of SyntEO. Here, the ontology describes the generation of an RGB remote sensing scene in 3D space, the following technical description of the pipeline uses parameters that are defined in the ontology.

\begin{figure}[H]
\centering
\includegraphics[width=16 cm]{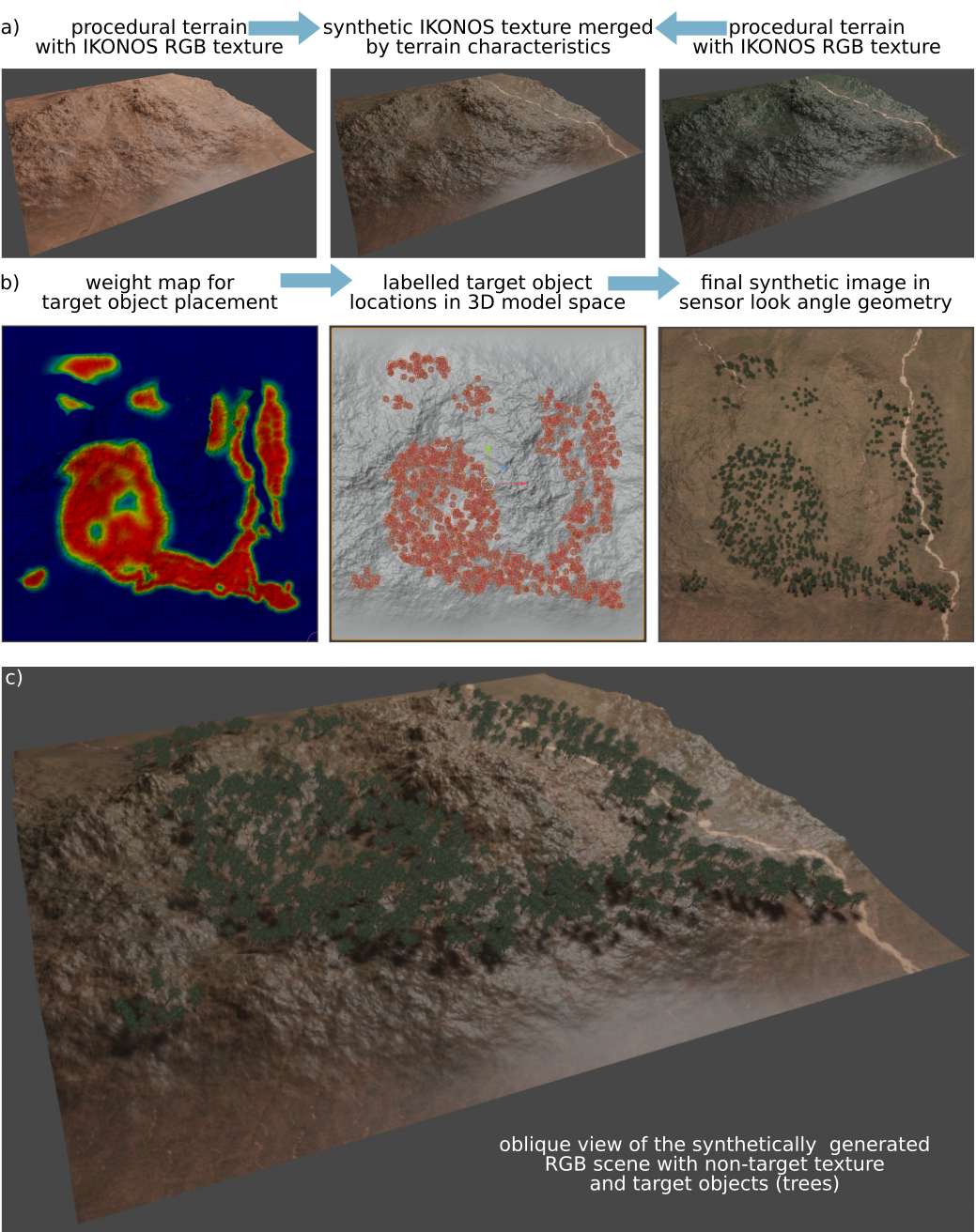}
\caption{Procedural generation of a high resolution RGB scene in Blender. a) IKONOS texture is merged based on characteristics of the generated terrain. b) Tree placement by employing a weight map and particle system; the final virtual remote sensing image from an overhead perspective. c) oblique view of the generated scene in 3D space.}
\label{fig:rgb}
\end{figure}

Figure \ref{fig:rgb} shows how the open source software Blender \cite{blender2021blender} is used to generate a complex high resolution RGB remote sensing scene. Initially, a procedural terrain is generated by using Perlin noise \cite{perlin1985animage}. Two crops from atmospherically corrected and pansharped IKONOS RGB images with a spatial resolution of 80~cm are used as texture and mixed using characteristics from the underlying procedural terrain. For demonstration, the target objects in this example are trees. Therefore a 3D model of a tree is randomly changed and distributed on the procedural terrain. For tree placement, a particle system uses a weight map that relies on terrain properties like height, surface convexity and pointiness, and a fixed offset to the edges of the scene extent to simulate a reasonable distribution. The output is an overhead image of the scene to simulate the sensor look angle.

The lightning of a synthetic scene can also be controlled to simulate shadow properties, which might be important in cities with higher buildings or in areas with steep terrain. The tree labels are derived from the particle system, which provides coordinates for each tree location and a radius depending on the randomly selected scale of the 3D parent object.

SyntEO should be understood as a system to structure expert knowledge with the goal to make it machine readable and inject it to a processing backend which then can compose synthetic scenes and finally provide a synthetically generated remote sensing image with corresponding labels. New tasks, sensors or drastically changing environmental conditions need to update the formalised expert knowledge and possibly a new processing backend. However, the longer SyntEO is in use, the more descriptions following the SyntEO ontology will exist and therewith, building new scenes will become faster and easier over time by fusing already used functionalities with new developments.

\end{multicols}
\begin{multicols}{2}

\section{Offshore Wind Farm Detection with SyntEO}

Offshore wind farms (OWF) are a crucial cornerstone for carbon-neutral strategies, which will be realised in the upcoming decades. For example, the European Union is planning to increase its annual energy generation with OWFs from today 12~GW to at least 60~GW in 2030 and 300~GW in 2050 with a financial investment of EUR~800 billion \cite{ec2020owf}. However, OWFs are on the interface of climate protection, human intervention in sensitive ecosystems, economic interests and state legislative \cite{gusatu2021spatial}. In order to support maritime spatial planning, the development of new OWF sites which are aligned to exclusion zones for natural conservation or fishing industry, the monitoring of OWF deployment and future studies which investigate the impacts of large scale OWFs, an independent, high resolution, spatial-temporal monitoring of OWFs is necessary. Since the deployment of OWFs is a global issue, the monitoring has to be globally too. Thus, Earth observation imagery is the most promising data source to provide a global, independent, constant and automatic OWF monitoring product in time \cite{zhang2021global}. To process a global data set frequently, efficient processing is indispensable, and at the same time, the model's spatial transferability must be guaranteed to avoid high false positive or false negative rates.

In this introduction to the SyntEO approach, a first hands-on example is demonstrated: The detection of offshore wind farms on two of the worlds largest offshore wind energy production sites, the North Sea basin and the East Chinese Sea. Furthermore, the model's performance on two additional non-OWF sites is taken into account to demonstrate the sensitivity for false detections in challenging environments and the ability of SyntEO to create large-scale data sets that can be used to train spatial transferable deep learning models.

\subsection{Studysite and Data}

Four study sites were selected due to their different characteristics to investigate the capabilities of the SyntEO approach. Most of the OWFs in the North Sea Basin are located with considerable distance to the coast and few of them are under construction. In comparison, OWFs in the East China Sea show different characteristics. Especially in the Hangzhou Bay near Shanghai or areas with strong tidal influence like the Jiangsu Rudong Offshore Intertidal Demonstration Wind Farm in the south of Jiangsu, OWFs appear in highly dynamic, near coast environments with many OWFs under construction. Hence the East China Sea is considered to be more challenging than the North Sea Basin concerning OWF detection. The other two study sites are the Persian Gulf and the Sea of Azov, both selected due to potentially high false positive rates resulting from offshore rigs for oil and methane extraction or rectangular agriculture fields with a specific size and geometric patterns, respectively.

The remote sensing imagery used during the SyntEO workflow and for prediction of the four study sites is based on C-band synthetical aperture radar (SAR) from Sentinel-1. Specifically, a stack of all acquisitions in three months from July to September 2020 of VH polarised IW GRD products from both ascending and descending orbits is reduced to a single band of median values. Furthermore, instead of the measurements in decibel, which are available as 16~bit floating-point numbers, the radiometric resolution is downscaled to 8~bit and the range of the values rescaled to a range of 0 to 255. Even when this results in a loss of information in each pixel, the spatial representation is only insignificantly influenced. Therewith, we follow a core idea of SyntEO that single pixel values are less important than spatial patterns, which are still distinct in 8 bit and a spatial resolution of 10 m $\times$ 10~m of the final S1-median product.

\begin{figure}[H]
\centering
\includegraphics[width=16 cm]{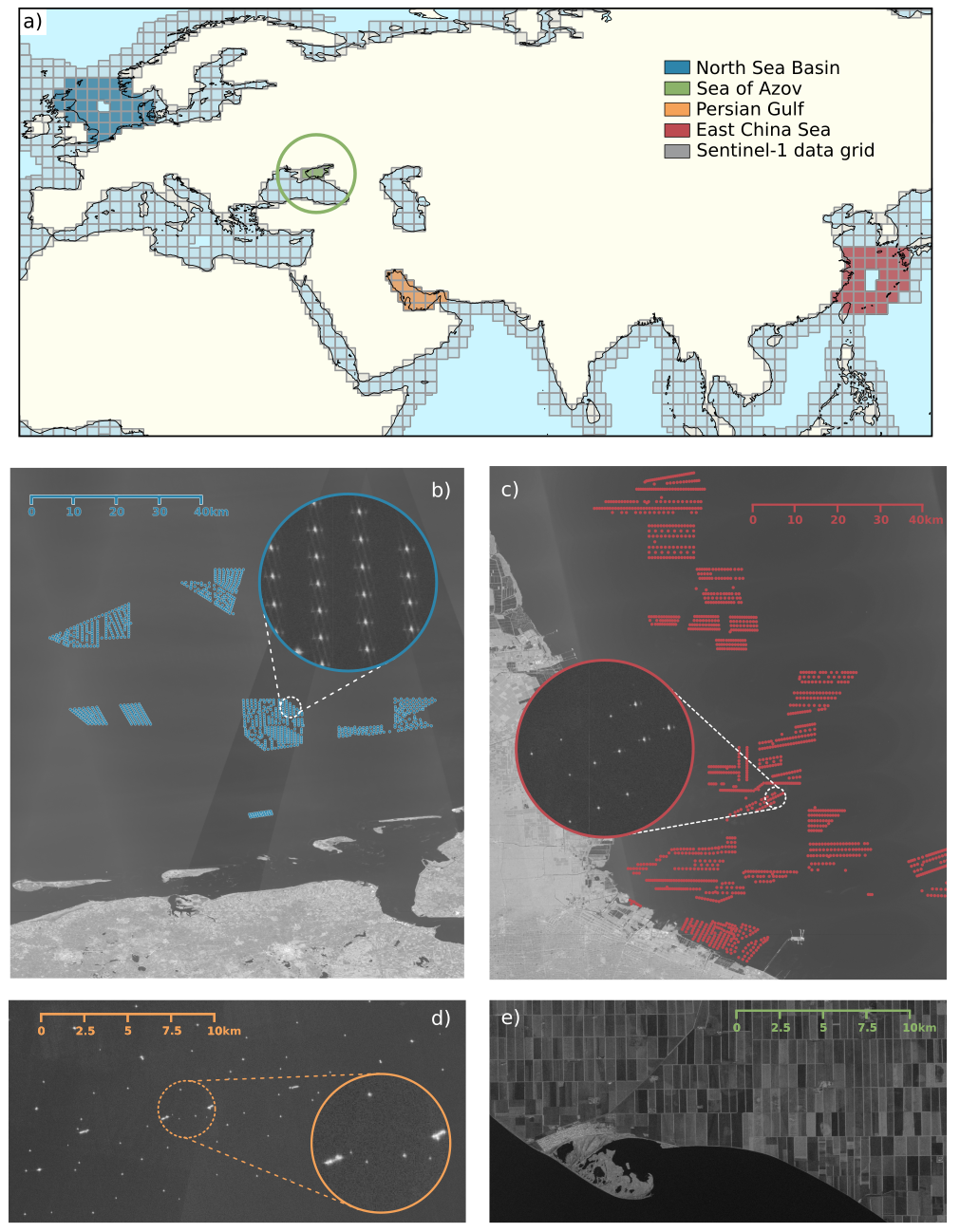}
\caption{Studysite location with examples of S-1 median images in 2020Q3. a) Locations of the four study sites and the defined S-1 data grid of the global coastline. b) Closeup of the North Sea Basin and OWFs in the German Bight. c) Closeup of the East China Sea, OWFs at the coast of Jiangsu, China. d) Closeup of the Persian Gulf, rigs in open water near Khaji, Saudi Arabia. e) Closeup of the Sea of Azov, agriculture fields used as false positive examples due to their grid like pattern near Yeysk, Russia.}
\label{fig:stusi}
\end{figure}

GoogleEarthEngine (GEE) \cite{gorelick2017gee} was used to query the preprocessed Sentinel-1 collection and create the 8~bit median stacks. In order to handle the query and download process on the GEE a 200~km buffer around the global coastline provided by OSM was used as area of interest. This polygon was structured in a database with tiles of 1.8° $\times$ 1.8°, see figure \ref{fig:stusi}. The finally downloaded Sentinel-1 median global data set for the third quarter of 2020 was split into two groups, one with potential OWF locations, like the mentioned areas above and another with scenes showing a mixture of land, coast and the open sea. The second group will later be used as texture template data in the SyntEO workflow.

For the North Sea Basin and the East China Sea, all OWFs were manually labelled to generate a ground truth test data set for the final evaluation of the SyntEO performances. Therefore, all completely deployed offshore wind turbines until the third quarter of 2020 were labelled with points. Upon these points, clusters were aggregated to receive the ground truth bounding boxes of offshore wind farms. In the North Sea Basin, this resulted in 3,787 offshore wind turbines within 42 OFWs and in the East China Sea, 1,587 offshore wind turbines in 25 OWFs.

\end{multicols}
\begin{multicols}{2}

\subsection{SyntEO Offshore Wind Farm Data Set Generation}

For a demonstration of SyntEO in a real-world application, a synthetic data set for OWF detection in multi-temporal Sentinel-1 images is created. In order to get the best impression of the creation and performance of the final data set, three preliminary data sets are forked during the data set generation. Thus, four consecutive synthetic data sets are produced with increasing complexity of the included synthetic training examples. For each data set, full from-scratch training of a well established CNN object detector, the Faster R-CNN \cite{he2016deepres} with adjacent performance analyses on the ground truth data set is carried out.

When looking at the OWFs shown in figure \ref{fig:stusi} the received perception can be formulated as follow: "\textit{Offshore wind farms are located in the sea but can appear in coastal areas on tidal flats. Smaller wind farms are closer to the coast than larger wind farms. The wind turbine density decreases by increasing OWF size. Wind turbines are organised in a regular grid-like pattern with individual but consistent, systematic changes to the grid structure for each wind farm. The typical outer shape of the entire wind farm is a polygon with four to five sides}."

With the above given semantic description, an expert can now formulate an ontology to structure this perceptual knowledge and translate it to numeric knowledge to make it machine-readable. For a better intuition, this example will focus on the target scene element \textit{WindFarm} as an example of how the formulated ontology can be used to generate this specific scene element.

\begin{landscape}

\begin{figure}
\centering
\includegraphics[width=25 cm]{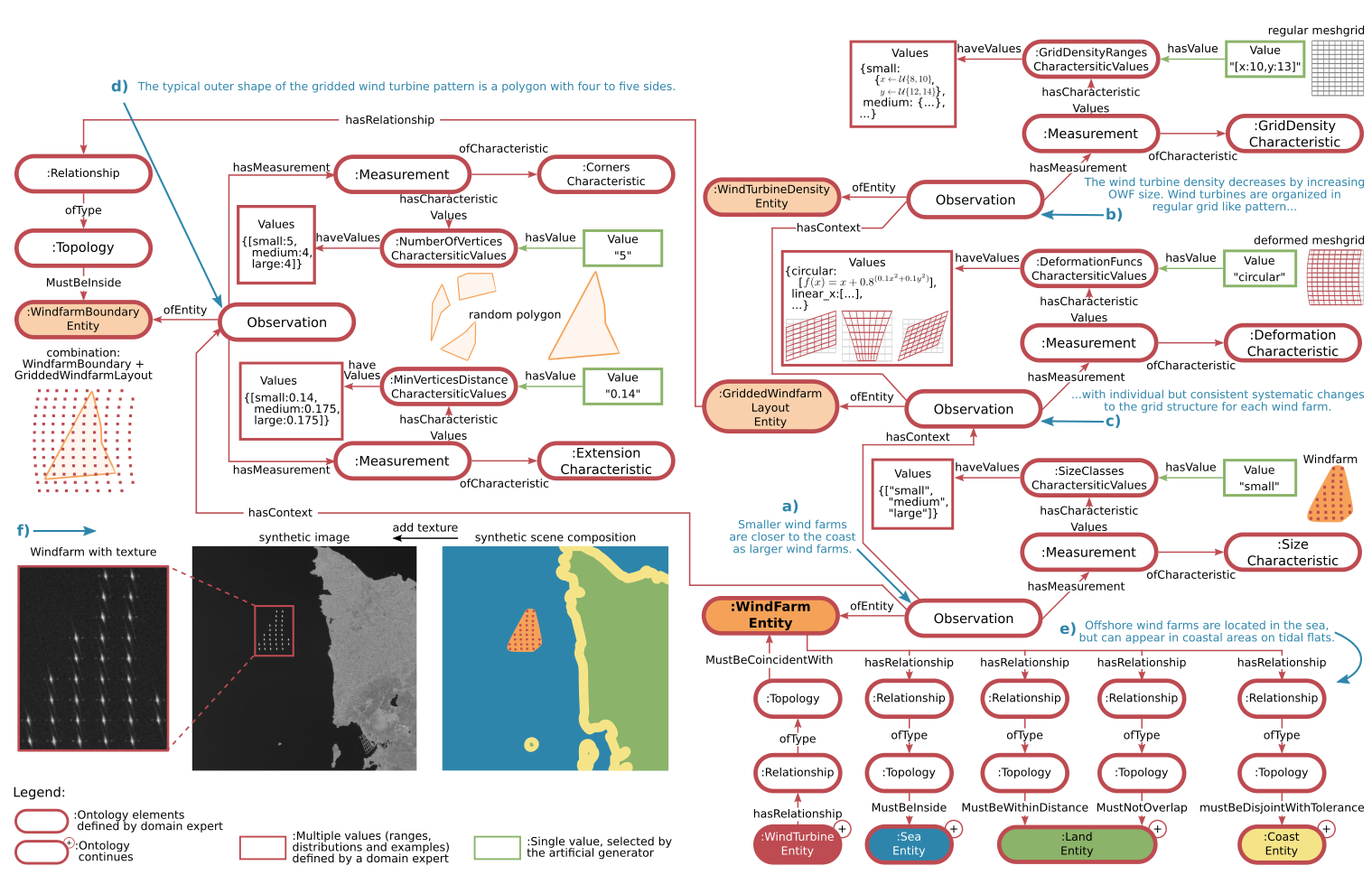}
\caption{Overview of the subset for \textit{WindFarm} generation of the SyntEO ontology for OWF data set generation. A semantic description of perceptual knowledge in blue leads through the ontology outlined. a) describes the size selection for the wind farm; b) and c) define the internal structure of the wind farm by potential turbine locations; d) describes the outer boundary of the wind farm and combines it with the potential turbine locations; e) relates the generated wind farm geometry to other scene elements by using topologies, resulting in the scene composition, f) shows how the scene composition is filled with texture and the derived bounding box annotation.}
\label{fig:ex_ontology}
\end{figure}

\end{landscape}

A graphical representation of the subset of \textit{WindFarm} related ontology is shown in figure \ref{fig:ex_ontology}. When following the structure of the ontology, the above given semantic description is incorporated in the visualisation of the ontology to show which point of the ontology represents the perceptual knowledge. The starting point of the present scenario is a scene composition with a predefined maximum scene extent of 20,480~m $\times$ 20,480~m. The artificial data generator has already included the non-target scene elements \textit{Sea, Coast} and \textit{Land} to the scene composition by using the ontology. The following task is to generate an OWF and integrate it into the existing scene composition.

Starting at a) in figure \ref{fig:ex_ontology}, first, the size of the OWF is taken from three values (large, medium and small), which represent scale factors as numeric values as well as semantics for better human understanding and to use the words as keys for subsequent choices of contextual \textit{Observations}. Since the \textit{Entity Land} is within distance of potential OWF locations in the scene composition, the data generator selects the \textit{Value} \textit{small:5 km} for the \textit{WindFarm Characteristic size}. After that, the artificial data generator generates the internal structure and outer boundary of the OWF. In b), the internal structure is modelled as a regular grid, where each crossing is a potential wind turbine location. The density of the grid is the number of lines in a unit square. This number is chosen randomly upon two discrete uniform distributions for the x and y-axis. The limits of these distributions are determined by the \textit{size Value} to represent the increasing wind turbine density with decreasing OWF size. The result is a regular, orthogonal grid in a unit square. To introduce c) "\textit{individual but consistent, systematic changes to the grid structure}", a deformation function is chosen randomly and applied to the generated regular grid.

In d), the outer boundary is defined to be a polygon of a specific number of vertices and distance to each other. Like the turbine density, the outer boundary \textit{hasContext} with the \textit{size} of the OWF. By using the \textit{Value small} as key, numeric values for \textit{NumberOfVertices} and \textit{MinVerticesDistance} are selected and passed to a random polygon generator, which again uses a unit square to construct the polygon as the outer boundary. Due to the \textit{Relationship ofType Topology} which determines that the \textit{GriddedWindfarmLayout MustBeInside WindfarmBoundary}, the potential wind turbine locations are spatially subsetted. The locations selected in this way finally constitute the internal and external structure of the OWF.

To further integrate the generated OWF into the scene composition, it is scaled by the initially selected value \textit{small:5 km}. Now in e), the \textit{Relationship ofType Topology} to Sea, Land and Coast can be used to find a location where all topologies are valid. Upon the final scene composition, the texture is added to receive the synthetic image in f). Therefore two approaches are applied. Sea, coast and land texture are queried from a template database. Since the geometries of these three entities rely on OSM coastline data, the same geometries can be used to query the prepared Sentinel-1 data grid of the median image. Thus, the shapes are filled with the corresponding pixel values.

On the other hand, the texture for wind turbines is completely procedural. The parameters for its automatic generation are also defined in the ontology in the \textit{Entity WindTurbine}. The generator module is a numpy array implementation that uses these parameters to create a stack of 2D kernels to generate each wind turbine's texture. In order to guarantee smooth gradients between sea and turbine texture, the values of the turbine texture are generated after the sea texture was added. This way, the local sea texture can be taken into account during procedural turbine texture generation. Therewith, geometries for scene composition and the corresponding texture are a combination of template data and procedural data provided by modules that ingest the input parameters from the ontology to compose and harmonise scene elements in the scene composition and texture adding process.

\end{multicols}
\begin{multicols}{2}

Finally, the corresponding label for the generated image, a bounding box of the OWF, is derived from the \textit{WindFarm} scene element and its final location in the scene composition. The image is exported to a single band .png file, whereas the generated label is saved to a .xml file following the PASCAL-VOC convention for object detection annotation \cite{everingham2010pascal}.

\begin{figure}[H]
\centering
\includegraphics[width=14 cm]{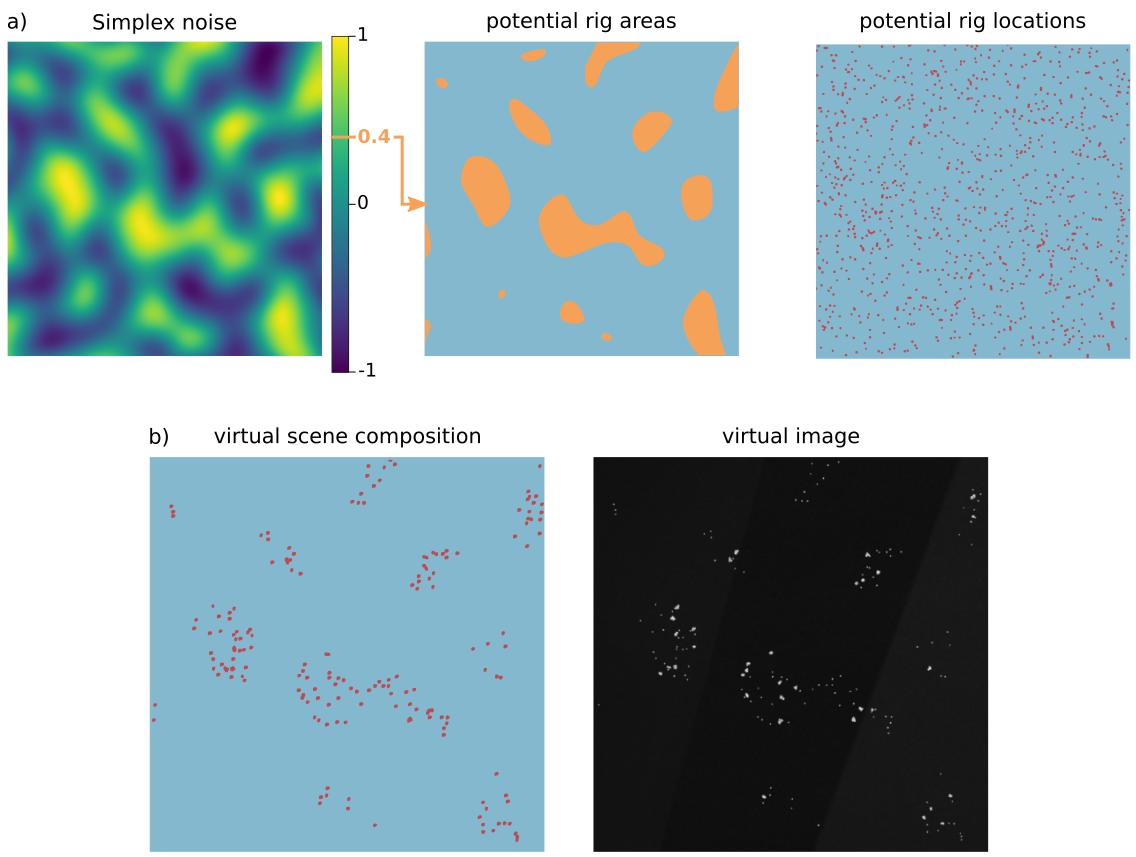}
\caption{Overview of the generation process of synthetic oil rig training examples. With a threshold on Simplex noise, generic shapes are generated. Their combination wit random locations for oil rigs is then filled with texture to generate the final non-target training example.}
\label{fig:rigs}
\end{figure}

With this introduced ontology, a data set can be created that only contains examples that show target information. However, it is good praxis in a large deep learning data set to include non-target examples, especially when non-targets exist that are known to be similar to the targets. In the case of OWFs these are, for example, oil rigs. As shown in figure \ref{fig:stusi}, single oil rigs are similar to single wind turbines, bright spots in front of a dark sea. However, the critical difference is that on a small scale, a single oil rig looks different from an offshore wind turbine and on a larger scale, oil rig clusters appear unstructured and random, whereas wind turbines in an OWF are structured. A description of oil rigs was also included in the ontology. To give an intuition about the generation process, figure \ref{fig:rigs} describes the procedural generation of oil rigs.
\end{multicols}
\begin{multicols}{2}
Therefore, simplex noise \cite{perlin2001noise} generates a 2D raster with values of ranges of [-1,1]. By applying a threshold to this raster, random organic shapes are created. Another module generates random locations with x and y coordinates, drawn from a uniform distribution in the same synthetic scene extent as the simplex noise. The rig locations are set up by applying a topology that only allows locations within the generated polygons. For adding texture to the rigs, the same approach as for the turbines is used. A 2D kernel simulates a single rig, whereby the kernel for a rig is a single Gaussian variant. Sea texture is again template data selected from the median Sentinel-1 data grid.

That way, non-target training images can be generated. However, the corresponding label file has to be created as specifically empty. During training, the optimiser recognises that explicitly no target is in the image and therewith, each prediction made during training by the model results in a false positive optimisation signal.

This demonstration of the SyntEO workflow shows how the perception by observation of real-world examples can be used to formulate a semantic description which is further developed to be an explicit description and finally machine-readable numeric knowledge. With the core classes of the SyntEO ontology, a structure can be created which, on one hand, allows to define \textit{Dimensions} of \textit{Values} to describe many \textit{Obeservations} and on the other hand, to give access to an artificial data generator which follows the same structure to generically compose complex scene compositions by selecting a single \textit{Value} for each \textit{Characteristic} it has to express.

\subsection{Synthetic Data Set Evolution}

With the previously described workflow, four sequential data sets were created. This was done to demonstrate the human-machine interaction that is possible by using the SyntEO approach. It shows how different synthetic data sets and changes to their generating parameters trigger different training results in a deep learning model. That way, an optimal ontology can be formulated by iterating through a dialogue between the human domain expert and the machine learning model.

\begin{figure}
\centering
\includegraphics[width=16 cm]{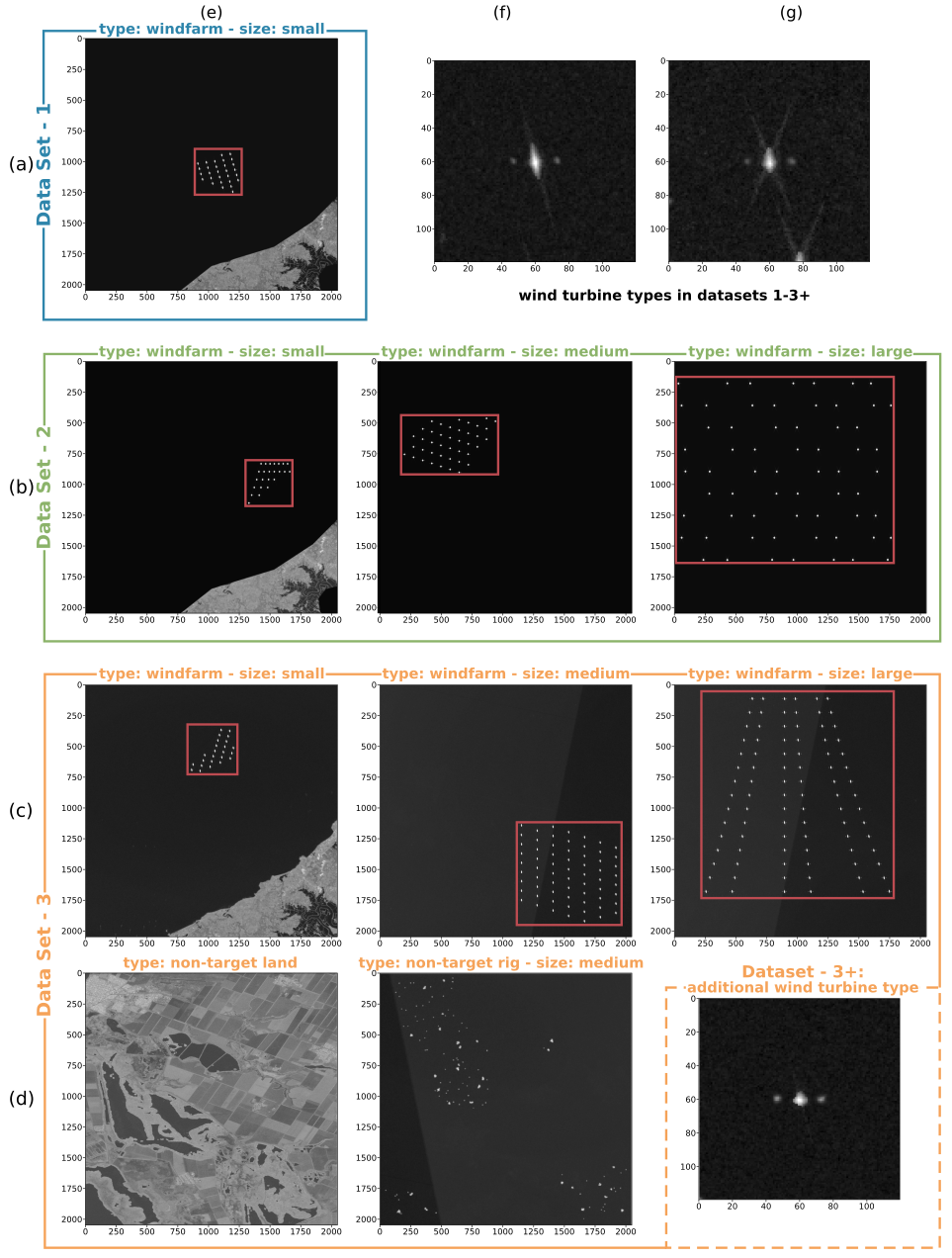}
\caption{Overview of the sequential synthetic data set evolution. From single sized (1) to multi-sized (2) OWFs and finally multi-sized OWFs in complex environments with template data for sea, land and coast plus non-target examples to minimise false detections (3,4). (1bc) show examples of procedural wind turbine textures used in data set 1-3, (4c) is a specific texture example for turbines on tidal flats, which are introduced in data set-3+.}
\label{fig:ds_evo}
\end{figure}

 The successive data sets each share the characteristics of the previous one and progressively increase in complexity. The data sets 1-3 and 3+, as shown in figure \ref{fig:ds_evo}, are created by stepwise enabling parts of the formulated ontology. Data set-1 shows only OWF of small size and near a land scene element without a coast scene element between land and sea. Also, the sea texture is not template data from the median Sentinel-1 data grid but a single constant value drawn from a uniform distribution. The overall size of the data set is 45,000 images. The handling of land without coast and a single value for sea data is maintained in data set-2. The changes for data set-2 were made by introducing the two additional size scales medium and large, to the OWF scene element. The overall size of data set-2 is 90,000 images with a 1/4-2/4-1/4 split for the size classes small, medium and large, respectively.
\end{multicols}
\begin{multicols}{2}
Data set-3 shifts the focus from target characteristics to non-target characteristics to make a model trained on the data set sensible for false positives. In data set-3, the non-target scene element coast is included, and texture data for the scene element sea is no longer a single value but template data from the median Sentinel-1 data grid. Furthermore, specific non-target examples are generated which show no OWF but scene elements that, on different scales, share specific characteristics with the target OWF. These non-targets are scenes with rectangular fields and road networks that show a regular gird like pattern generated with template data. The other are examples that show rigs for e.g. oil and methane extraction as irregular clusters with a diameter of 200~m to 20~km of bright spots in front of darker sea texture.

The last data set-3+ has the same scene elements and texture as data set-3 but an additional texture for a specific offshore wind turbine type. In the data sets 1-3 two turbine texture types were used, see figure \ref{fig:ds_evo} 1b) and 1c). Both types show wind turbines in the open sea, whereas the X like pattern of 1c) is found in regions with a balanced number of Sentinel-1 acquisitions of ascending and descending orbits, like in the North Sea Basin. Similar numbers of acquisitions from both orbits are necessary for this particular pattern, which originates from Doppler effects that are clearly visible over open water. Whereas in the East China Sea, the 1b) version is more frequently found, which means acquisitions of one orbit dominate the median signal. However, the 4c) variant specially designed for data set-3+ shows a specific type of wind turbine that normally occurs near the coast in areas with a strong tidal influence. Due to the tidal flats, some acquisitions in the three-month temporal stack have images without water. These acquisitions dampen the occurrence of the Doppler effect to be visible in the image since over the tidal flats, the radar backscatter properties drastically change compared to open water. Therewith, the unique fingerprint of a typical offshore wind turbine is less distinct for examples in such areas. To represent these special cases, they were included in the data set-3+. For both data sets, 3 and 3+, the overall size is 90,000 training examples with a split of 1/6-1/3-1/6-1/6-1/6 for the OWF sizes small, medium and large and the non-targets rigs and land, respectively.

\subsection{CNN Model and Training}

The Faster R-CNN object detector \cite{ren2015faster} with a ResNet-50 \cite{he2016deepres} convolutional feature extractor was used as the only CNN architecture in each training. This specifically basic and well-established object detector in Earth observation was chosen to demonstrate that with a suitable data set, basic deep learning architectures can achieve sound results on the ground truth data. Therewith, in this study, we purposely shift the focus away from details in variations of architectures and training schemes towards different variants of the training data set. Thus the ability of SyntEO to support stable experiment environments by having control over the training data set becomes more obvious in this demonstration.

The only difference in the Faster R-CNN configuration between the different trainings is the initial anchor size and ratio of the Region Proposal Network (RPN), which is characteristic of the Faster R-CNN architecture. Briefly described, the RPN is a submodule of the Faster R-CNN, which connects the ResNet-50 feature extractor with the Faster R-CNN object detector head. The RPN uses fixed anchor boxes to check for each anchor if it contains a potential object. Thereby, the anchor boxes can be imagined as uniformly distributed boxes over the last feature map that comes from the ResNet-50 backbone. Since potential objects might occur on different scales and with different aspects, for each anchor box location, $n$ anchor boxes are processed with different scales and aspect-ratios \cite{ren2015faster}. Normally, the number of boxes per anchor is defined by an expert due to an analysis of the training data set. In the case of SyntEO, the necessary information can directly be derived by consulting the ontology and synthetic image extent.

In this example, the maximum scene extent is 20,480~m $\times$ 20,480~m, with a synthetic sensor resolution of 10~m, equal to the ground sampling distance of Sentinel-1 IW GRD products, each synthetic image has a dimension of 2048 $\times$ 2048 pixels. The available hardware for this study is a NVIDIA RTX-2080Ti GPU with 11~GB memory. Since training on the full resolution of 2048 $\times$ 2048 pixels would result in a memory overflow, the earlier discussed second variant of defining the synthetic image extent for training was chosen. The reason is that large OWFs often have a dimension of 17~km in diameter and should appear in a single training example. Nevertheless, since 2048 $\times$ 2048 pixels are too large for the given hardware constraints, the images are getting downscaled before training and therewith accepting the loss of small scale features. Finally, the image extent of a training example and the CNN model's input dimension are 1024 $\times$ 1024 pixels, which cover the full synthetic scene extent of 20,480~m $\times$ 20,480~m.

By looking into the ontology, three maximum [5~km, 10~km, 17~km] and two minimum [1200~m, 2400~m] sizes can be extracted as major OWF extents. With a spatial resolution of 10~m $\times$ 10~m and to allow easier calculation in this example, the scale values get slightly adjusted to [128, 256, 512, 1024, 1792]. Since these pixel dimensions describe the size of the targets in the synthetic image but not in the feature map which enters the RPN, the downscaling factor of the ResNet-50 backbone, the stride, must be taken into account. For the given ResNet-50 architecture, the stride is 16. Furthermore, the size of a scale one anchor box is set to 4 $\times$ 4 pixels in this example. Now, the following equation can be used to calculate scale factors for each object size derived from the ontology:

\begin{equation}\label{eq:scale}
\mathcal{A}_{\mathrm{scale}}=\sqrt{\mathcal{T}_{h}\mathcal{T}_{w}\frac{\mathcal{M}_{h}\mathcal{M}_{w}}{\mathcal{I}_{h}\mathcal{I}_{w}}}\times\frac{1}{s\sqrt{\mathcal{A}_{h}\mathcal{A}_{w}}}
\end{equation}

Where $\mathcal{A}_{\mathrm{scale}}$ is the anchor box scale factor, $_h$ and $_w$ are height and width in pixel, $\mathcal{T}$ is the size of a target object, $\mathcal{M}$ the input size of the CNN model, $\mathcal{I}$ the size of the synthetic image, $s$ the stride of the CNN feature extractor and $\mathcal{A}_{h}\mathcal{A}_{w}$ height and width of the anchor box of scale 1. By applying (\ref{eq:scale}) for the above given target object sizes, data set-1 has the scale factors [0.25, 0.5, 1] and all other data sets have the scale factors [0.25, 0.5, 1, 2, 3.5], the aspect ratios are same for all [0.5, 1, 2].

The training was done on four parallel NVIDIA RTX-2080Ti GPUs by using the NVIDIA docker implementation of the TensorFlow deep learning framework. The .png images and .xml annotation files were parsed to the TFRecord binary format to enable TensorFlow's data API. The images in the TFRecord files have the size of 2048 $\times$ 2048 pixels and are downscaled to an input size of 1024 $\times$ 1024 pixels on the fly during training. Eight training shards in TFRecord format hold 95\% of all examples, whereas the single validation shard contains the remaining 5\% for each synthetic data set. The test data set is not part of the synthetic data set but is the independent, manually labelled ground truth data set of the four introduced test sites. Further hyperparameter settings during training are a momentum optimiser with a momentum of 0.9 and a learning rate with a cosine decay \cite{loshchilov2017sgdr} with a base of 0.01 over 10,687 or 21,375 training steps and a batch size of 4 for data set-1 and all other data sets, respectively.

\subsection{Test Site Prediction and Metrics}

After training, each model was exported to predict the ground truth (GT) data set. Therefore, the GT tiles were split into chips of size 2048 $\times$ 2048 with a 50\% overlap, which are again resized to 1024 $\times$ 1024 during prediction. A non-maximum suppression with a threshold of 0.8 on the predicted score was applied for all bounding boxes. Since some OWF clusters are very close in the ground truth data and spread over multiple image chips, the remaining bounding boxes were aggregated by converting their coordinates to the geographic reference system WGS84 and merging them in a cascading manner by starting with the highest prediction score and a minimum IoU of 0.333, see (\ref{eq:iou}), with the next bounding box. That way, an accurate description of the estimated location and shape by the trained model over multiple images on the ground truth data is maintained. The polygons created by this approach are the final result of the test data. They are saved as .geojson file and used to calculate metrics for the accuracy assessment and model performance investigation.

For model evaluation and comparison, the following metrics where calculated:

\begin{equation}\label{eq:iou}
\textrm{IoU}=\frac{\textrm{area}(y_{\textrm{gt}} \cap y_{\textrm{pred}})}{\textrm{area}( y_{\textrm{gt}} \cup y_{\textrm{pred}})}
\end{equation}

Upon the IoU, the threshold $\tau$ of 0.33 is applied to define whether a prediction is true positive (TP) or false positive (FP). The threshold reflects the technical difference of ground truth bounding boxes and predicted shapes. Where predictions are tight polygons with irregular shapes around the OWFs, ground truth are rectangular bounding boxes which were automatically derived from the single turbine point clusters. In dynamic areas like the East China Sea, these bounding boxes cover a significantly larger area than the predicted polygons. Thus, a lower IoU is better suited to balance the different shapes of predictions and labels. Now, the performance can be measured by precision Pr, recall Rc and F1 score:

\begin{equation}
\mathrm{Pr}=\frac{\mathrm{TP}}{\mathrm{TP+FP}}
\end{equation}

\begin{equation}
\mathrm{Rc}=\frac{\mathrm{TP}}{\mathrm{TP+FN}}
\end{equation}

\begin{equation}
\mathrm{F1}=2 \times \frac{\mathrm{Pr}\times \mathrm{Rc}}{\mathrm{Pr}+\mathrm{Rc}}
\end{equation}

To calculate the average precision AP a monotonic precision-recall curve, $\mathrm{Pr}_\mathrm{interp}$ is constructed by sorting all predictions indexed with $k = 1,2,...,K$ in descending order of their score. For each $k$ the subsequent precision-recall pair is calculated. Thereby, $\mathrm{Pr}_\mathrm{interp}$ always corresponds to the next maximum in the ordered recall values, by using the all-point interpolation approach, see Padilla~\etal \cite{padilla2021metrics} for a detailed explanation:

\begin{equation}
\mathrm{Pr}_\mathrm{interp}=\underset{\tilde{\mathrm{Rc}}\colon\tilde{\mathrm{Rc}}\ge \mathrm{Rc}}{\mathrm{max}}\ \mathrm{Pr}_\mathrm{interp}(\tilde{\mathrm{Rc}})
\end{equation}

In order to summarise the characteristics of the precision-recall curve in a single metric, the average precision AP is reported by calculating the Riemann integral of the precision-recall curve:

\begin{equation}
\mathrm{AP}=\sum\limits_{k=1}^{K}({\mathrm{Rc}}(k)-{\mathrm{Rc}}(k-1))\times \mathrm{Pr}_\mathrm{interp}(\mathrm{Rc}(k)),
\end{equation}

with $\mathrm{Rc}(0)=0$.

\section{Results}

The result section focuses on presenting the performance of SyntEO in the real world example OWF detection to give a good intuition on how SyntEO can be used in order to investigate the training process and model behaviour. An in-depth discussion about SyntEO itself is given in section \ref{sec:discussion}.

\end{multicols}
\begin{figure}[H]
	\centering
	\includegraphics[width=12 cm]{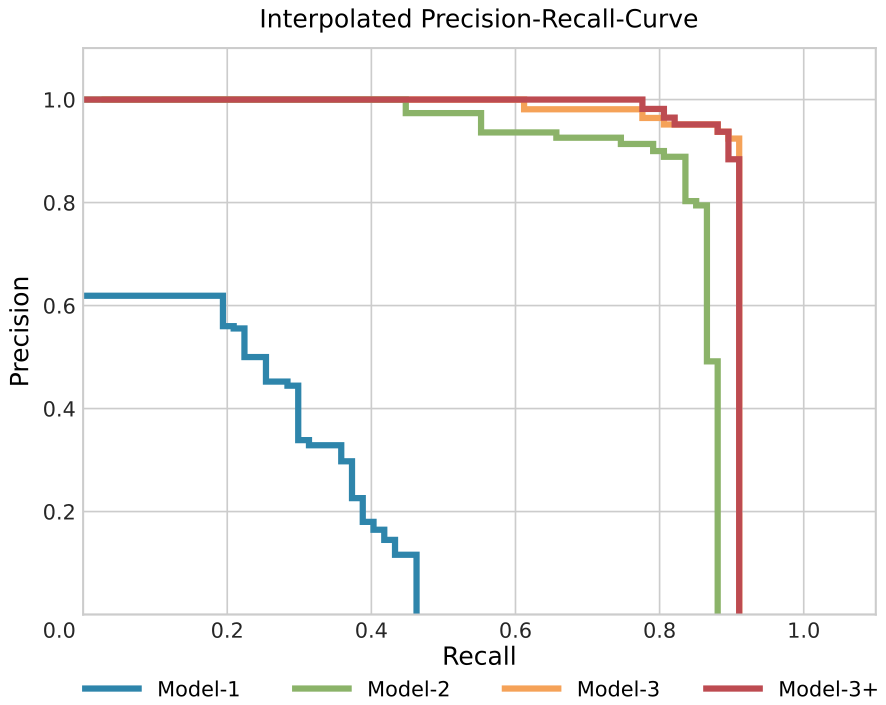}
	\caption{Interpolated precision-recall curves for the four trained models and their performance on the ground truth data set. The corresponding AP values are: Model-1 (0.21), Model-2 (0.842), Model-3 (0.901), Model-3+ (0.904).}
	\label{fig:prcurve}
\end{figure}

\begin{table}[H]
\footnotesize
\caption{Overview of all metrics for all models on the ground truth (GT) data. For each model, the metrics are presented \textbf{Combined} for all test sites together and separated for each test site. For the \textbf{Combined} performance, the \underline{best metrics are underlined}. True positive (TP); false positive (FP); false negative (FN); recall (Rc), precision (Pr); average precision (AP). Metrics with $_{\mathrm{WT}}$ refer to the model performance on a wind turbine level.}
\label{tab:metrics}
\centering
\begin{tabular}{llllllllllllll}
\toprule
\textbf{Studysite}&GT&TP&FP&FN&Rc&Pr&F1&AP&{}&GT$_{\textrm{WT}}$&TP$_{\textrm{WT}}$&$\mathrm{Rc}_{\mathrm{WT}}$\\
\midrule
\textbf{Model - 1}&{}&{}&{}&{}&{}&{}&{}&{}&{}&{}&{}&{}\\
\textbf{Combined}&\textbf{67}&\textbf{31}&\textbf{252}&\textbf{36}&\textbf{0.463}&\textbf{0.11}&\textbf{0.177}&\textbf{0.21}&{}&\textbf{5374}&\textbf{863}&\textbf{0.161}\\
{North Sea Basin}&{42}&{19}&{85}&{23}&{0.452}&{0.183}&{0.260}&{0.217}&{}&{3787}&{687}&{0.181}\\
{East Chinese Sea}&{25}&{12}&{67}&{13}&{0.48}&{0.152}&{0.231}&{0.282}&{}&{1587}&{176}&{0.111}\\
{Persian Gulf}&{0}&{}&{100}&{}&{}&{}&{}&{}&{}&{}&{}&{}\\
{Sea of Azov}&{0}&{}&{0}&{}&{}&{}&{}&{}&{}&{}&{}&{}\\
\midrule
\textbf{Model - 2}&{}&{}&{}&{}&{}&{}&{}&{}&{}&{}&{}&{}\\
\textbf{Combined}&\textbf{67}&\textbf{59}&\textbf{80}&\textbf{8}&\textbf{0.881}&\textbf{0.424}&\textbf{0.573}&\textbf{0.842}&{}&\textbf{5374}&\textbf{5263}&\textbf{0.979}\\
{North Sea Basin}&{42}&{40}&{3}&{2}&{0.952}&{0.93}&{0.941}&{0.952}&{}&{3787}&{3764}&{0.994}\\
{East Chinese Sea}&{25}&{19}&{22}&{6}&{0.76}&{0.46}&{0.576}&{0.712}&{}&{1587}&{1499}&{0.945}\\
{Persian Gulf}&{0}&{}&{31}&{}&{}&{}&{}&{}&{}&{}&{}&{}\\
{Sea of Azov}&{0}&{}&{24}&{}&{}&{}&{}&{}&{}&{}&{}&{}\\
\midrule
\textbf{Model - 3}&{}&{}&{}&{}&{}&{}&{}&{}&{}&{}&{}&{}&{}\\
\textbf{Combined}&\textbf{67}&\underline{\textbf{61}}&\underline{\textbf{11}}&\underline{\textbf{6}}&\underline{\textbf{0.91}}&\underline{\textbf{0.847}}&\underline{\textbf{0.878}}&\textbf{0.901}&{}&\textbf{5374}&\textbf{5185}&\textbf{0.965}\\
{North Sea Basin}&{42}&{40}&{1}&{2}&{0.952}&{0.976}&{0.964}&{0.952}&{}&{3787}&{3742}&{0.988}\\
{East Chinese Sea}&{25}&{21}&{5}&{4}&{0.84}&{0.808}&{0.824}&{0.817}&{}&{1587}&{1443}&{0.91}\\
{Persian Gulf}&{0}&{}&{5}&{}&{}&{}&{}&{}&{}&{}&{}&{}\\
{Sea of Azov}&{0}&{}&{0}&{}&{}&{}&{}&{}&{}&{}&{}&{}\\
\midrule
\textbf{Model - 3+}&{}&{}&{}&{}&{}&{}&{}&{}&{}&{}&{}&{}&{}\\
\textbf{Combined}&\textbf{67}&\underline{\textbf{61}}&\textbf{14}&\underline{\textbf{6}}&\underline{\textbf{0.91}}&\textbf{0.813}&\textbf{0.86}&\underline{\textbf{0.904}}&{}&\textbf{5374}&\underline{\textbf{5296}}&\underline{\textbf{0.985}}\\
{North Sea Basin}&{42}&{40}&{0}&{2}&{0.952}&{1}&{0.976}&{0.952}&{}&{3787}&{3756}&{0.992}\\
{East Chinese Sea}&{25}&{21}&{4}&{4}&{0.84}&{0.84}&{0.84}&{0.831}&{}&{1587}&{1540}&{0.97}\\
{Persian Gulf}&{0}&{}&{10}&{}&{}&{}&{}&{}&{}&{}&{}&{}\\
{Sea of Azov}&{0}&{}&{0}&{}&{}&{}&{}&{}&{}&{}&{}&{}\\
\bottomrule
\end{tabular}
\end{table}
\begin{multicols}{2}
Four data sets were generated with a minimum size of 45,000 and a maximum size of 90,000 training examples. For the largest data sets with 90,000 examples, it took a machine with 4 Intel Xeon Platinum 8260 CPUs and 2.40GHz running 190 parallel threads, 2.6 hours to generate all images along with their annotation files. Therewith, the SyntEO approach for this example shows the potential of how a large training data set can be generated on demand once an ontology is formulated and connected to an image processing backend.

All four synthetic data sets were used to successfully optimise four deep learning models, which will further be called model-1 to 3+, corresponding to the data set. Due to the increasing complexity of the data sets, models 3 and 3+ are trained upon the most representative data sets, which results in the best performance metrics see figure \ref{fig:prcurve} and table \ref{tab:metrics}. On a wind farm level, both models achieve recall scores of 91\%. On a wind turbine level, the performance increases to 96.5\% and 98.5\% for models 3 and 3+, respectively. That demonstrates that the models are able to detect most wind farms confidently, and the predicted boundaries enclose almost all wind turbines of the ground truth data. At the same time, the precision values of model 3 and 3+ are well above over 80\%, resulting in high F1 scores greater than 0.85 and an AP greater 0.9. Therewith, the models are not just able to securely detect OWFs but at the same time to minimise the false detection rate.

\begin{figure}[H]
\centering
\includegraphics[width=16 cm]{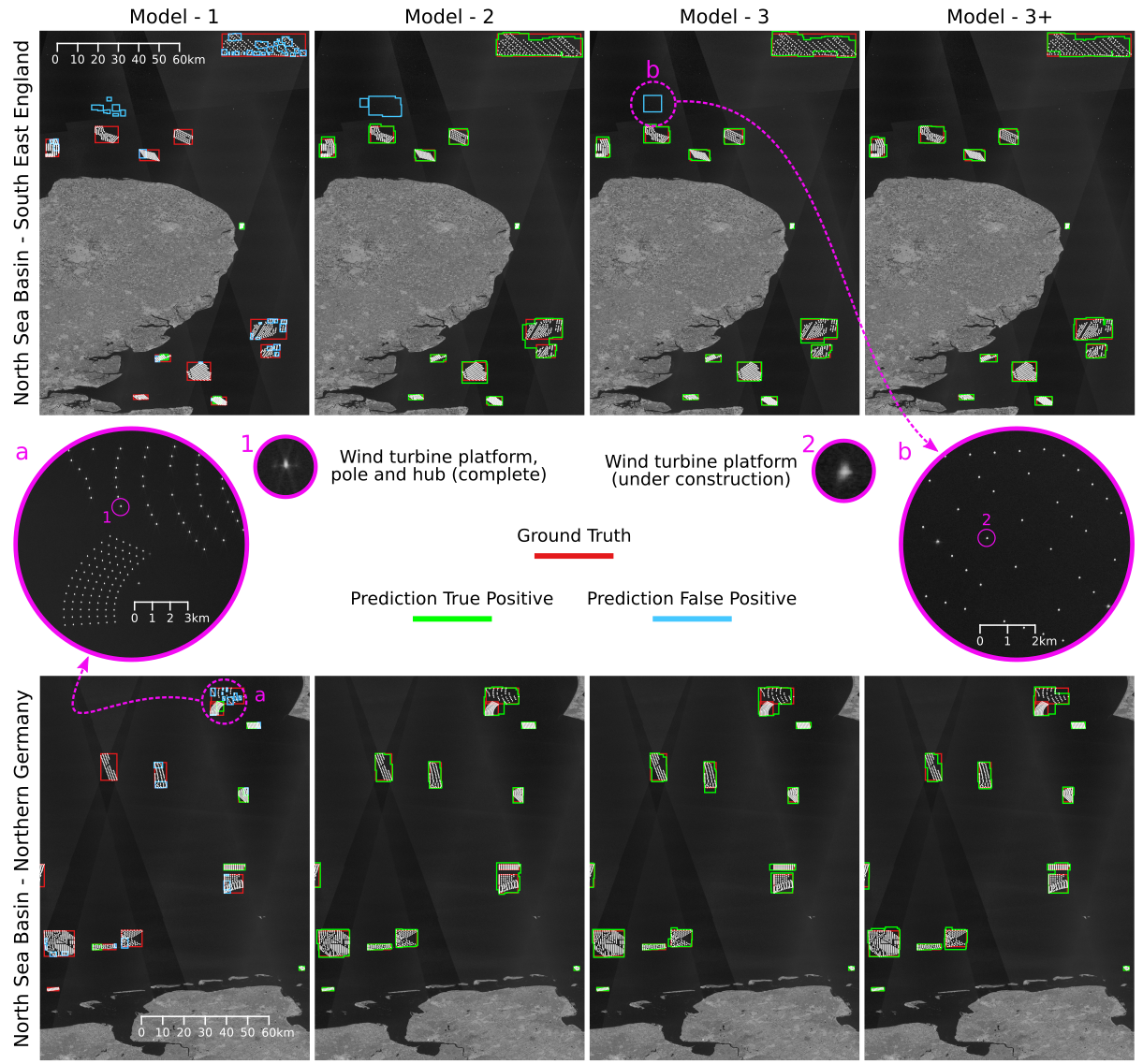}
\caption{Model performances on the North Sea basin test site with two examples in South East England and Northern Germany. Closeup a) shows the two OWFs Horns Rev 2 and 3; closeup b) shows the Triton Knoll OWF which in 2020Q3 was under construction.}
\label{fig:res_nsb}
\end{figure}

\begin{figure}[H]
\centering
\includegraphics[width=16 cm]{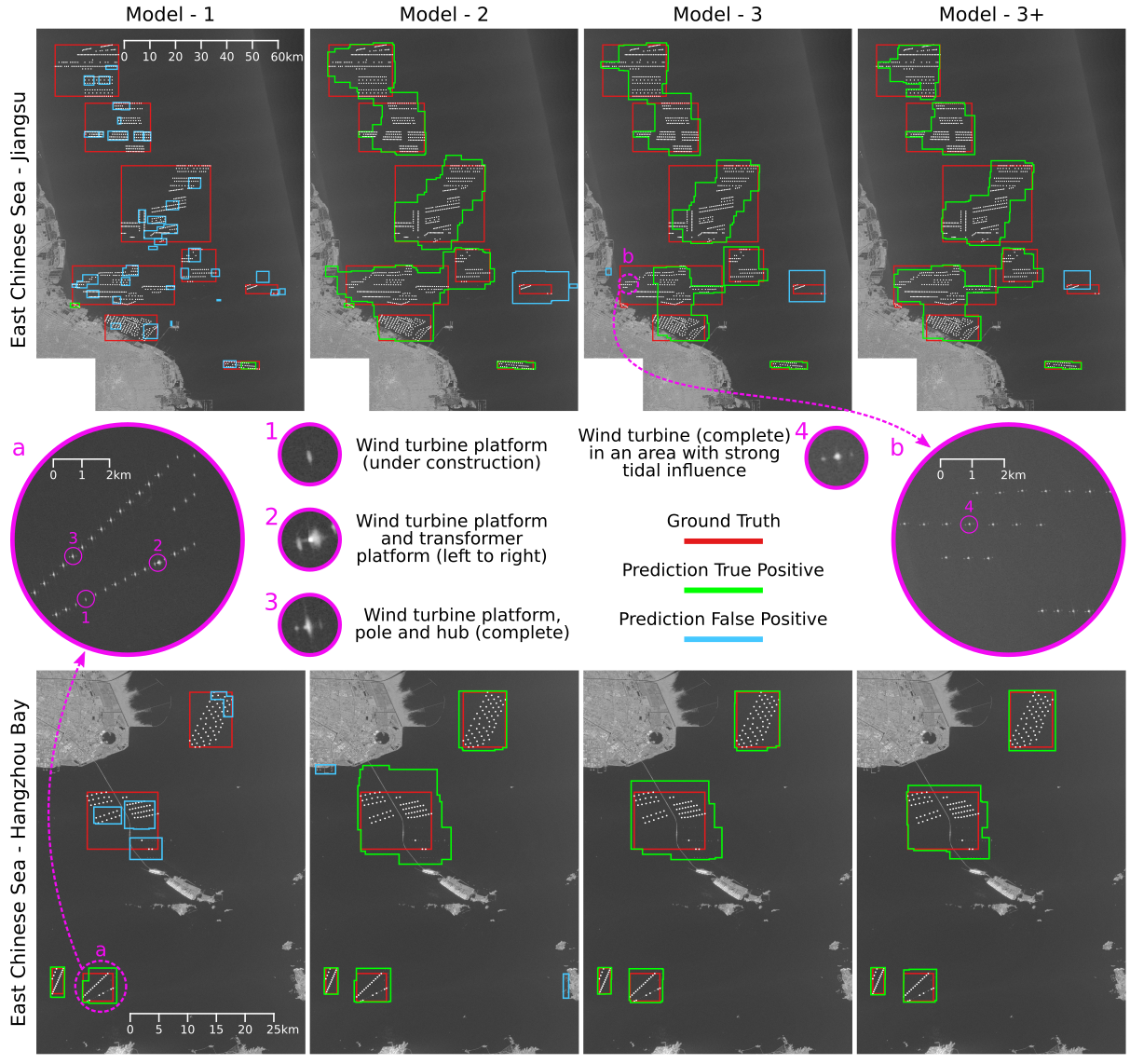}
\caption{Model performances on the East Chinese Sea test site with two examples in the south of Jiangsu and the Hangzhou Bay area near Shanghai. Closeup a) shows an OWF, which is still under construction but with the most turbines completely deployed; closeup b) shows an OWF situated on tidal flats with a less distinct wind turbine signature in the Sentinel-1 median image.}
\label{fig:res_ecs}
\end{figure}

Figure \ref{fig:res_nsb} shows examples of the North Sea Basin test site. The progression in model performance is clearly visible in increasing boundary refinement of the predicted OWF areas and eventually no false detection. The introduction of all size parameters in data set-2 strongly influences the model performance of model-2, resulting in less fragmented predictions than model-1, since model-2 learns to make predictions on multiple scales. The false positive detection in the South East England example, which remains until model-3, is the Triton Knoll OWF which in 2020Q3 was under construction. Magnification b2) shows a platform's spatial pattern under construction without turbine pole and hub compared to magnification a1) of a completely deployed wind turbine. The detection is a false positive since the ground truth and synthetic training data only contain completely deployed wind turbines. The structure on a medium to large scale is already similar to a completed OWF, making it a challenging example. Interestingly, the predicted area decreases from model-2 to 3 after introducing the non-target oil rigs, which on the smallest scale look similar to a wind turbine platform under construction.
\end{multicols}
\begin{multicols}{2}
In comparison to the North Sea Basin, the entire East China Sea test site is more complex, see figure \ref{fig:res_ecs}. Many OWFs are under construction with less distinct boundaries and in less homogenous areas, surrounded by harbour infrastructure, small islands and bridges. Furthermore, fewer acquisitions lead to less distinct wind turbine signatures in the S1-median images, see magnification a3). The general trend of increasing boundary refinement over all models and the onset of multi-scale detection from model-2 is the same as in the North Sea Basin test site. However, the Hangzhou Bay example of model-2 shows that when the model is trained to look on larger scales, pier structures are getting confused for OWFs. By introducing the coast scene element in data set-3, model-3 is aware of all sizes of OWF targets and pier structures as non-targets. This way, the model is now optimised to differentiate between regular structures coming from piers and OWFs. Along with the introduction of the non-target coast in data set-3, oil rigs were also introduced in this data set. Even when oil rigs do not appear in the East China Sea, these non-target training images are the reason that in the Jiangsu example, model-3 underestimates the western part of the near coast OWF cluster, see magnification b). The near coast OWF in Jiangsu is built on tidal flats, resulting in less distinct turbine signatures, as discussed earlier, see magnification 4b). These specific turbines are more similar to wind turbines under construction or oil rigs. Since model-3 has learned to differentiate between oil rigs and OWFs, these wind turbines are wrongly rejected. Interestingly, model-3 securely detects the OWF cluster beneath that area, closer to the coast and under the same tidal influence. The difference between both clusters is that the lower cluster, closer to the coast, has a stronger grid-like pattern on a medium scale, since this wind farm, the Jiangsu Rudong Offshore Intertidal Demonstration Wind Farm, is completely deployed. This observation of model behaviour supports the hypothesis that model-3 looks at features from small and medium scales. Furthermore, model-3 puts a stronger weight on a regular pattern in the medium scale as on single wind turbine features on a small scale and therefore predicts correctly on the near coast cluster and false on the less structured cluster above. Also, this is a possible explanation why model-3 wrongly includes the unfinished Triton Knoll OWF in the North Sea Basin. Finally, by including the specific wind turbine texture for turbines in tidal areas in data set-3+, the corresponding model-3+ is able to detect the western part of the OWF cluster in the Jiangsu test site.
\end{multicols}

\begin{figure}[H]
\centering
\includegraphics[width=16 cm]{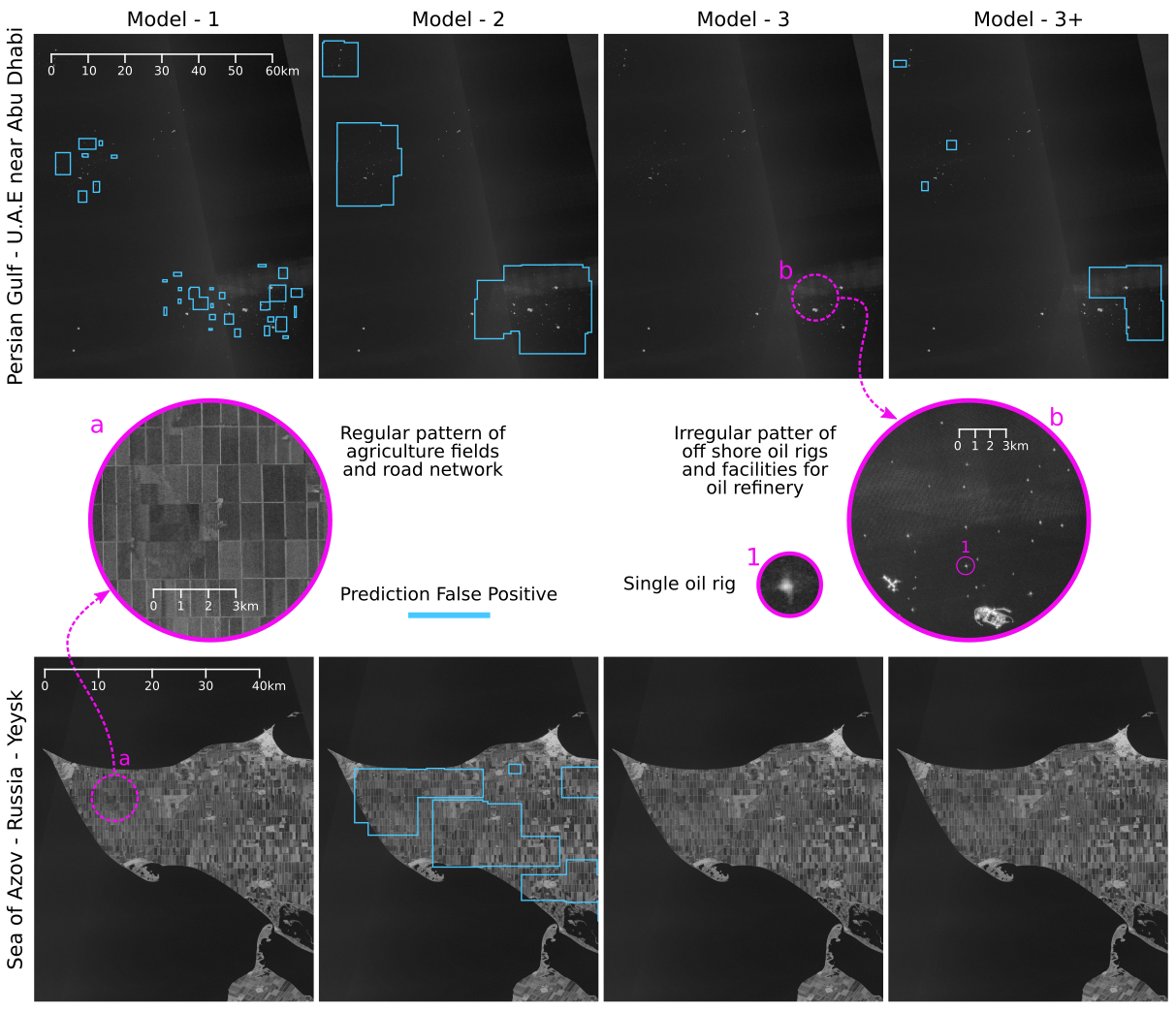}
\caption{Model performances on the Persian Gulf and Sea of Azov test sites with challenging non-targets. Closeup a) shows rectangular gridded patterns of agricultural fields and road network on the coast of the Sea of Azov; closeup b) shows oil rigs and refineries in the Persian Gulf.}
\label{fig:res_pg_soa}
\end{figure}
\begin{multicols}{2}

Figure \ref{fig:res_pg_soa} shows the two non-target test sites. The Persian Gulf example shows how model-1 and 2 are unaware of non-targets similar to OWFs, since they are not included in the training data set. As in the OWF test sites, the size adaption is clearly visible from model-1 to 2. With the introduction of synthetic oil rigs as non-target training examples, the model dramatically reduces its false positive rate. However, with the introduction of the wind turbine texture for strong tidal areas, model-3+ starts to detect false positives again since the small scale feature of target class OWF is now again closer to oil rigs as in model-3. The Sea of Azov example shows how, after introducing the OWF typical grid-like patterns in medium and large scales, model-2 detects similar patterns on a similar scale within field and road networks. With the introduction of non-target inland training examples in data set-3, the models 3 and 3+ can differentiate between field-road network patterns and OWFs securely and reduce the false detection rate to 0.

Overall, model-3 and model-3+ are the best performing model variants, which are trained with data sets including both: a high variation in targets as well as a high variation in non-targets. This combination makes both models spatially transferable. That was demonstrated in reliable detection of OWFs at two OWF test sites, which have different characteristics in the underlying data, with less optimal conditions in the East China Sea, as well as target and non-target complexity. Furthermore, the demonstrated approach can reduce false positives by specifically including non-target training examples in the synthetic training data and thereby dramatically affecting the precision of the model performance. Besides the control over the model behaviour, the SyntEO approach also supports gathering insights in the investigated target in real-world applications. That way, the turbine type in strongly tidal areas could specifically be addressed by increasing wind turbine texture variance. Therewith, the human-machine interaction, which the domain expert started with the formulation of the ontology, continues. The model response can be interpreted as an answer that suggests that a specific anomaly in the target data was not taken into account by the human expert.

\section{Discussion}\label{sec:discussion}

The ontology in SyntEO is a complex set of rules for a representative description of the investigated object and its common context. Hence, one could ask why this set of rules is not employed in a rule-based classification. Furthermore, the core concepts of SyntEO like the H-resolution model \cite{strahler1986on}, hierarchically nested scene elements \cite{wu1995hierarchy} and the usage of auxiliary data coming from the GIS domain to investigate remote sensing images \cite{lang2006bridging} are together with a rule-based approach driven by expert knowledge characteristics of Geographic Object Base Image Analysis (GEOBIA) \cite{blaschke2010object, blaschke2014geographic, lang2019geobia}. Works from this field inspired the SyntEO approach. However, compared to a typical GEOBIA workflow, SyntEO does not rely on pre-segmentation before classification since feature extraction is part of the training that also optimises for classification and unifies this in a single model. Neither does it use thresholds in a rule-based approach directly, both of which are issues of GEOBIA and often mentioned why such approaches are less spatial transferable and partly rely on heavy fine-tuning by an expert \cite{arvor2019ontologies}. In order to avoid these limitations, in SyntEO, the expert knowledge is intensively used during data set generation and strongly influences the otherwise purely data-driven deep learning approach. The deep learning model, which is optimised upon a SyntEO data set, indirectly learns a combination of expert knowledge, auxiliary data, and underlying features coming from their final spatio-temporal composition. This way, the entire SyntEO workflow and final deep learning model training is a hybrid approach of expert knowledge-driven and data-driven image analysis. SyntEO can thus be classified as an approach on the interface between rule-based image analysis on one side and deep learning and Big-EO data on the other side.

During the demonstration of SyntEO, it could be shown how the approach can be utilised to establish human-machine interactions with fully controllable experiment environments and adjustable data set variants. Therewith, it was possible to make assumptions over the training process and model behaviour which is an important cornerstone towards explainable machine learning and interpretability of artificial intelligence, which is a current challenge in Earth observation and artificial intelligence \cite{tuia2021agenda}. The SyntEO approach starts a dialogue between machine and expert by using the ontology as a medium for exchange that both sides can interpret. Together with large scale ablation studies, there is the possibility of modelling complex systems and investigating them in order to explain the outcome of a prediction by comparing them with existing knowledge. Furthermore, such a stable and adjustable experiment environment can also be used to discover new insights like anomalies that diverge from a stable, defined training data set but are included and important characteristics in real-world data. In every application of SyntEO, the ontology serves as a basic structure that allows experts from different domains and backgrounds to set up an environment in which they can start a machine-human dialogue, regardless of whether the knowledge they bring to the ontology is based on physical models or has a purely artistic origin.

The aforementioned advantages of the SyntEO approach are in addition to the original motivation: The generation of large training data sets that are particularly suitable for Earth observation. The example given illustrates how the approach can generate a huge data set with up to 90,000 images, which is the multiple of existing real-world examples of the target objects. Due to the synthetic data set, it is now possible to train deep learning models in a supervised manner since the number of training examples is no longer an issue. Therewith, SyntEO provides the possibility to investigate objects which are scarce but scattered over many thousands of remote sensing images by applying a synthetic data approach. This opens up new opportunities for novel applications in Earth observation with deep learning.

However, an argument against a SyntEO data set might be that it will never be able to generate highly complex ecosystems or that the expenses are too high to build such a complex synthetic environment. The counterargument is that a hybrid approach of synthetic and hand labelled data should be considered before this red line is crossed. This way, one part of the training data is created with a less complex SyntEO approach, and another, smaller part of the training data would be annotated manually. During training, the SyntEO data set can be used as a pre-training data set to move the model's parameters in the right direction and finally transfer learning on the small but highly complex real-world data set to fine-tune the model and learn the real complexity of the given task \cite{yosinski2014transferable}.

\section{Conclusion}

By employing the SyntEO approach, it is possible to generate large scale, deep learning ready data set. SyntEO is based on fundamental characteristics of Earth observation data and strongly takes multi-scale, spatio-temporal context during data synthesising into account. Domain experts can include their knowledge in the data generation process by making it explicit and machine-readable by embedding it in an ontology. An artificial data generator uses this knowledge representation to build the training data set. Detailled control over the generation process can be gained by deactivating and activating parts of the ontology. Thus, it is possible to establish stable and highly adjustable experiment environments. Therewith, SyntEO offers the possibility to establish human-machine interactions which provide a fundament to gather insights in the machine learning process.

In a first hands-on example, offshore wind farms were detected in the North Sea Basin and East China Sea with a deep learning CNN model solely optimised on a SyntEO data set with the largest size of 90,000 generated images. High scores in recall and precision confirm the ability of the proposed approach to detect the target of interest by minimising false detections securely. Furthermore, conclusions about the training process and model behaviour could be made by conducting an ablation study with different data set variants. On a single local example of wind turbines in tidal flats, it was possible to show how the sound experiment design of SyntEO can be used to raise awareness of anomalies of the target object in real-world data, which are not included in the synthetic training data.

Overall, SyntEO is a flexible approach that is built around the introduced ontology concept, which makes expert knowledge accessible for automatic data generation in Earth observation. Future studies which apply the SyntEO approach will show how it can be utilised for different application domains to get new insights into Earth observation and deep learning.

\section*{Funding}
This research received no external funding.

\section*{Declaration of competing interest}
The authors declare no conflict of interest.

\section*{Author Contributions}
Thorsten Hoeser: Conceptualisation, original manuscript writing, code development, data processing, visualisation; Claudia Kuenzer: Supervision and manuscript reviewing.

{\small
\bibliographystyle{ieee_fullname}
\bibliography{synteo}
}

\end{multicols}

\end{document}